\documentclass[lettersize,journal]{IEEEtran}
\usepackage{amsmath,amsfonts}
\usepackage{algorithmic}
\usepackage{algorithm}
\usepackage{array}
\usepackage[caption=false,font=normalsize,labelfont=sf,textfont=sf]{subfig}
\usepackage{textcomp}
\usepackage{stfloats}
\usepackage{url}
\usepackage{verbatim}
\usepackage{graphicx}
\usepackage{cite}
\usepackage{mathbbol}
\usepackage{graphics} 
\usepackage{epsfig} 
\usepackage{amsmath} 
\usepackage{amssymb}  
\usepackage[breaklinks=true]{hyperref}
\usepackage{xcolor}
\usepackage{graphicx}
\usepackage{amsthm}
\newtheorem{thm}{Theorem}
\newtheorem*{rem}{Remark}

\begin{document}

\title{Mixed Reality Environment and High-Dimensional Continuification Control for Swarm Robotics}



\author{Gian Carlo Maffettone,~\IEEEmembership{Graduate Student Member,~IEEE,} Lorenzo Liguori, Eduardo Palermo,~\IEEEmembership{Member,~IEEE,} Mario di Bernardo,~\IEEEmembership{Fellow,~IEEE,}, Maurizio Porfiri,~\IEEEmembership{Fellow,~IEEE,}
\thanks{This work was developed with the economic support of MUR (Italian Ministry of University and Research) performing
the activities of the project PRIN 2022 “Machine-learning based control of complex multi-agent systems for search and rescue operations in natural
disasters (MENTOR) and of the National Science Foundation under Grant CMMI-1932187. (Mario di Bernardo and Maurizio Porfiri contributed equally.) (Corresponding authors: Mario di Bernardo, Maurizio Porfiri.)}
\thanks{Gian Carlo Maffettone and Mario di Bernardo are with the Modeling and Engineering Risk and Complexity program at the Scuola Superiore Meridionale, Naples, Italy (emails: giancarlo.maffettone@unina.it, mario.dibernardo@unina.it).}
\thanks{Lorenzo Liguori and Eduardo Palermo are with the Department of Mechanical and Aerospace Engineering, University of Rome La Sapienza, Roma (emails: lorenzo.liguori@uniroma1.it, eduardo.palermo@uniroma1.it)}
\thanks{Maurizio Porfiri is with the Dept. of Biomedical Engineering, Dept. of Mechanical and Aerospace Engineering, Tandon School of Engineering, New York University, New York, USA (email: mporfiri@nyu.edu).}
\thanks{Gian Carlo Maffettone, Lorenzo Liguori and Maurizio Porfiri are also with the  Center for Urban Science and Progress, Tandon School of Engineering, New York University, New York, USA.}
\thanks{Mario di Bernardo is also with the Dept. of Electrical Engineering and Information Technology, University of Naples Federico II, Naples, Italy.}
}

\markboth{}%
{}


\maketitle

\begin{abstract}
Many new methodologies for the control of large-scale multi-agent systems are based on macroscopic representations of the emerging system dynamics, in the form of continuum approximations of large ensembles. These techniques, that are developed in the limit case of an infinite number of agents, are usually validated only through numerical simulations. In this paper, we introduce a mixed reality set-up for testing swarm robotics techniques, focusing on the macroscopic collective motion of robotic swarms. This hybrid apparatus combines both real differential drive robots and virtual agents to create a heterogeneous swarm of tunable size. We also extend continuification-based control methods for swarms to higher dimensions, and assess experimentally their validity in the new platform. Our study demonstrates the effectiveness of the platform for conducting large-scale swarm robotics experiments, and it contributes new theoretical insights into control algorithms exploiting continuification approaches.
\end{abstract}


\section{Introduction} \label{sec:intro}
Several new techniques for the analysis and control of large-scale multi-agent systems rely on the assumption that the interacting dynamical systems of the ensemble (\textit{agents}) are numerous enough to  be described in a continuuum framework\cite{maffettone2022continuification, maffettone2023continuification, nikitin2021continuation, gao2019graphon, gao2023lqg}. Such an assumption paves the way for recasting many traditional \textit{microscopic} agent-based formulations, based on large sets of ordinary differential equations (ODEs), into  smaller sets of partial differential equations (PDEs) for a \textit{macroscopic} representation of their collective behavior. 
For instance, it can be advantageous to study the spatio-temporal dynamics of a large group of mobile agents in terms of their density, rather than keeping track of the motion of each of the agents \cite{maffettone2022continuification, maffettone2023continuification, nikitin2021continuation, bernoff2011primer, sinigaglia2022density}. In so doing, one can address the curse of dimensionality of microscopic representations by formulating control algorithms at the scale where the collective behavior emerges \cite{diBernardo2020}. Suitable applications include, but are not limited to, multi-robot systems \cite{freudenthaler2020pde, biswal2021decentralized, sinigaglia2022density}
, traffic control \cite{Karafyllis2022, blandin2011general}, cell populations \cite{rubio2022open}, 
and human networks \cite{calabrese2021spontaneous}.

Recasting these systems into continuum formulations offers new opportunities in the analysis and design of novel control approaches to tame collective dynamics. Pressing open challenges are: ($i$) to find agile methods to inform and experimentally validate the synthesis of  control algorithms developed in a continuum framework \cite{dorigo2021swarm}, and ($ii$) to design new strategies for controlling the collective behavior of large-scale multi-agent systems, such as those in swarm robotics, that exploit their macroscopic approximation in terms of PDEs \cite{d2023controlling}.

Regarding the first challenge, full-scale experiments about the control of large-scale multi-agent systems have been recently carried out \cite{slavkov2018morphogenesis, rubenstein2013collective, caprari2005mobile, mondada2009puck}. However, the majority of the existing control solutions have been tested only using computer simulations due to practical limitations. In this paper, we present a novel mixed reality environment where some real mobile robots interact among themselves and with other virtual agents. We bring settings as that in \cite{durham2011discrete} and other recent mixed reality platforms \cite{manas2023scalability, karunarathna2023mixed}, to large-scale scenarios. 
In so doing, we integrate insights from disability studies \cite{boldini2021virtual, ricci2023virtual} and animal behavior research \cite{naik2020animals, stowers2017virtual, karakaya2020behavioral, polverino2022ecology} where digital twins of patients or animals are often utilized for testing new strategies in virtual reality settings.
Our set-up let the user choose the size of the ensemble to study, avoiding the bottleneck of extreme time cost and resources of experiments of large-scale systems. Moreover, in our setting, the specific model for the virtual agents can be chosen by the designer and is not constrained to a specific commercial robot. The whole apparatus is easy to implement and can be realized, for example, by adapting other existing facilities such as the Robotarium at GeorgiaTech \cite{pickem2017robotarium}.
Relevant previous work in the field of swarm robotics includes the use of augmented reality for providing simple testbed agents, like kilobots, with augmented sensing capabilities \cite{reina2017ark, feola2023multi}.

\begin{figure}
     \centering
     \includegraphics[width=0.5\textwidth]{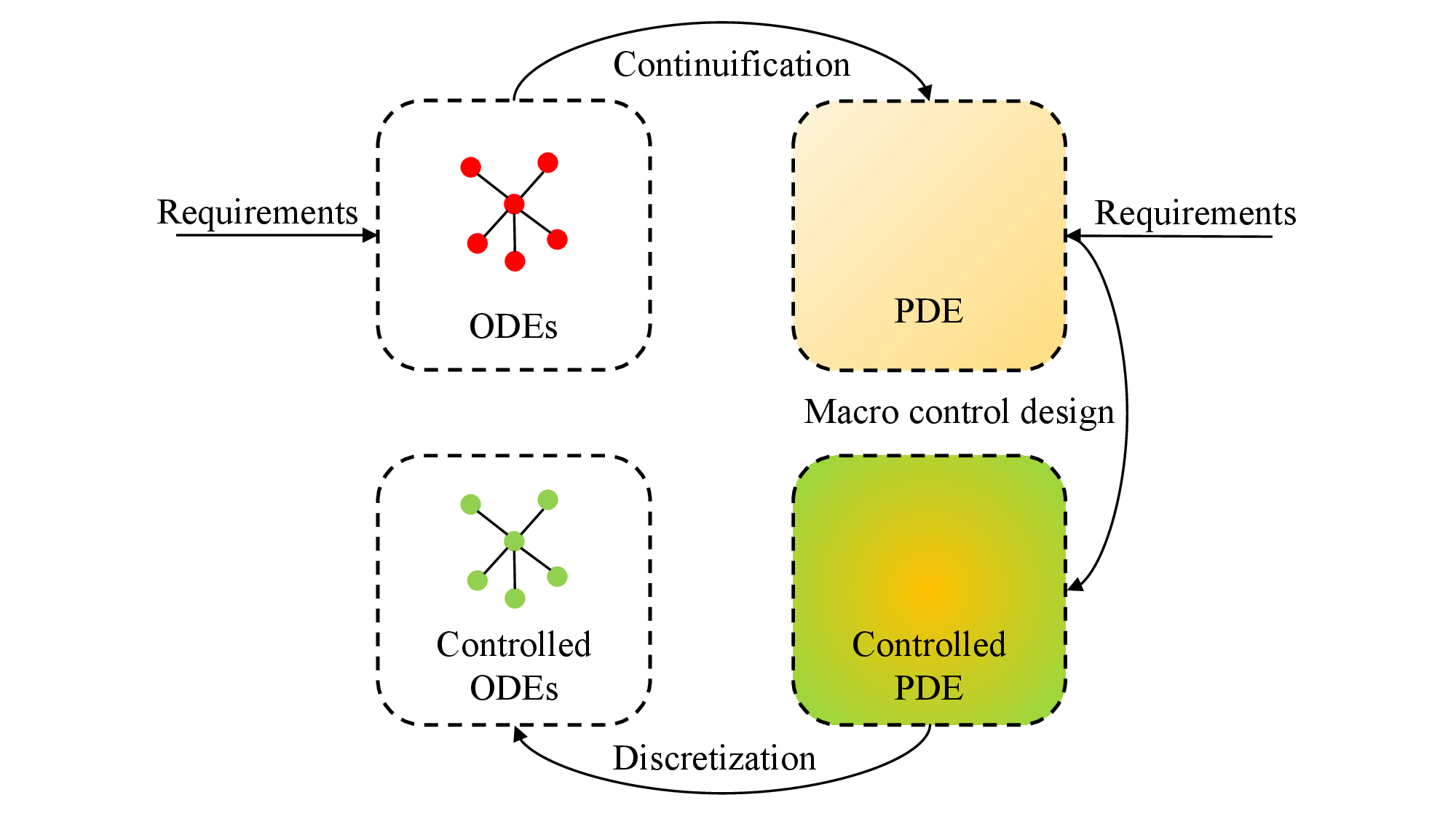}
     \caption{Continuification control scheme (inspired by \cite{nikitin2021continuation}). The schemes describes all the stages of the solution: ($i$) continuification, ($ii$) macroscopic control design, and ($iii$) discretization.} 
     \label{fig:continuification}
\end{figure}

To address the second challenge highlighted above, we address the theoretical problem of extending the continuification-based control approach presented in \cite{maffettone2022continuification, maffettone2023continuification} to higher-dimensions. Upon deriving a PDE describing the emergent collective behavior of the swarm we wish to control (continuification), we design a macroscopic control action that ensures convergence to a desired density. Such a control is then discretized to obtain deployable control inputs for the agents in the swarm (see Fig. \ref{fig:continuification}). We emphasize that the transition from the 1D case discussed in \cite{maffettone2022continuification} to the broader theoretical framework in higher dimensions is non-trivial from the technical viewpoint, as detailed in Section \ref{sec:control_design}. Specifically, during the discretization process, new degrees of freedom emerge and technical advancements are needed to ensure well posedness, and, eventually, fulfill additional control requirements.
Finally, the proposed platform is used to validate experimentally the theoretical framework that we developed.

The rest of the paper is organized as follows. 
In Section \ref{sec:exp_platform}, we describe the experimental platform. Specifically, in Section \ref{subsec:therobots} we focus on the mobile robots we designed, and then, in Section \ref{subsec:theplatform} on the platform itself. In Section \ref{sec:theoreticalframework}, we derive the theoretical extension to higher-dimensions of the continuification-based control approach that is validated experimentally in Section \ref{sec:demonstration} to demonstrate the use of the platform. We discuss results and conclusions in Section \ref{sec:discussion} and \ref{sec:conclusions}, respectively.

\section{Experimental mixed reality environment}\label{sec:exp_platform}
Here, we detail our experimental apparatus for the design of experiments about the coordination of hybrid large swarms of real robots and virtual agents. We first present the mobile robotic agents and their kinematics. Then, we describe the integration of these robots with the virtual agents in the overall mixed reality platform.

\subsection{Differential drive robots}\label{subsec:therobots}
\begin{figure}
\centering
\subfloat[]{\includegraphics[width=0.3\textwidth]{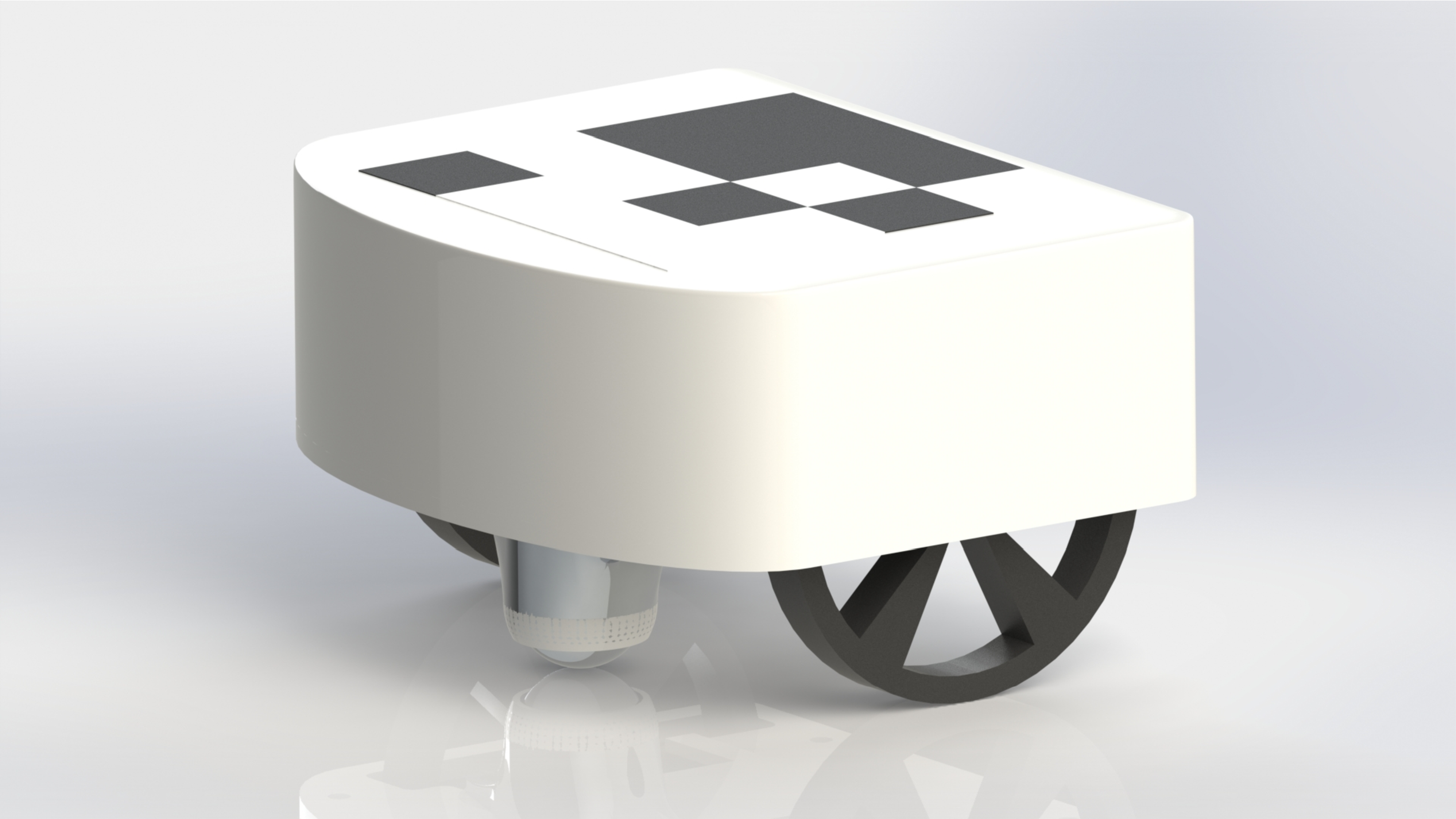}%
\label{subfig::robot}}
\hspace{0.1cm}
\subfloat[]{\includegraphics[width=0.4\textwidth]{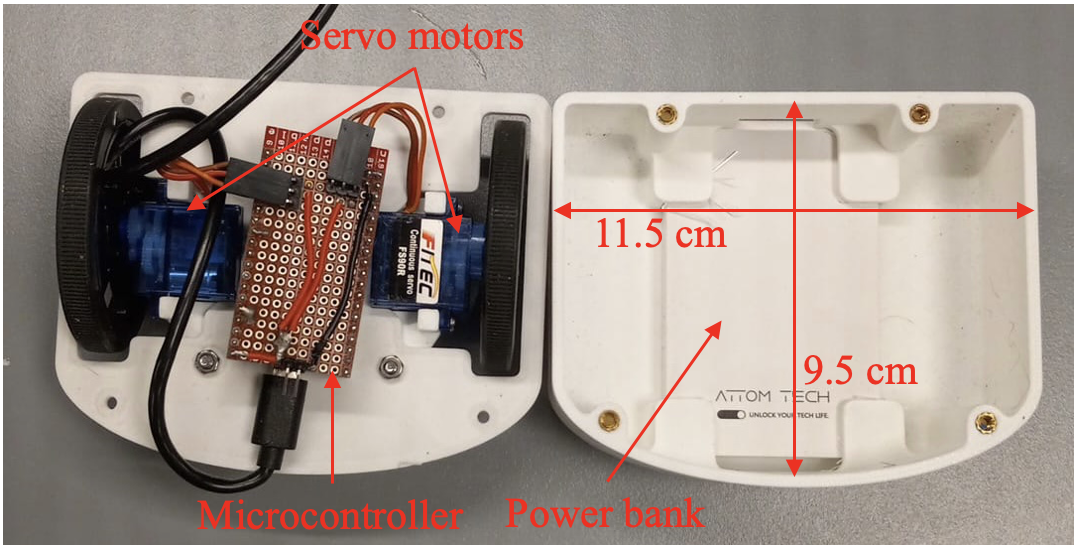}%
\label{subfig::openrobot}}
\caption{(a) Render of a differential drive robot, and (b) inner view of the robot.}
\label{fig:robots}
\end{figure}
We built four differential drive robots, as the one rendered in Fig. \ref{subfig::robot}. These robots featured a 3D-printed PLA frame (Polylite, Polymaker) printed on a Bambu Lab X1C (CAD model available at \href{https://github.com/Dynamical-Systems-Laboratory/ContinuificationControl}{\nolinkurl{https://github.com/Dynamical-Systems-Laboratory/ContinuificationControl}}). The sizes of the robot are such that it can be schematized as a rectangle $11.5\,\mathrm{cm} \times 9.5\,\mathrm{cm}$. Each robot was equipped with an ESP32 microcontroller, operating two continuous rotation servo motors (FS90R, Feetech) directly connected to 56 mm wheels.
Additionally, an omni-directional wheel was attached at the front-bottom of the robot. Power was supplied to each robot through an off-the-shelf power bank (Attom, Ultra Slim 3000mAh). We show the real robot, with sizes and hardware in Fig. \ref{subfig::openrobot}. In the absence of a load, the motors are able to rotate at approximately $14$ rad/s and provide a torque of 1.5 kg$\cdot$cm. Taking into account the wheel radius, the maximum linear speed that can be achieved by the robot is approximately 0.8 m/s (when both wheels are rotating in the same direction at full speed).

The $i$-th differential drive robots is characterized by the following non-holonomic kinematic behavior:
\begin{align}\label{eq:diff_drive_kinematic}
    \dot{\mathbf{z}}^\mathrm{R}_i = \mathbf{R}(\theta_i)\, \mathbf{u}_i^\mathrm{R},
\end{align}
for $i=1,\dots,4$. In particular, $\mathbf{z}_i^\mathrm{R} = [\mathbf{x}_i^\mathrm{R}, \theta_i]^T$ is the state of the $i$-th differential drive robot, where $\mathbf{x}_i^\mathrm{R}=[x^\mathrm{R}_{i,1}, x^\mathrm{R}_{i,2}]^T$ is its position in a Cartesian coordinate framework and $\theta_i \in [-\pi, \pi]$ its orientation. Moreover,
\begin{align}
\mathbf{R}(\theta)=
    \begin{bmatrix}
        \cos\theta_i & 0\\
        \sin\theta_i & 0\\
        0 & 1
    \end{bmatrix},
\end{align}
and $\mathbf{u}_i^R = [V_i, \omega_i]^T$ is the vector of the control variables,
with $V_i$ being the instantaneous velocity of the mid-point between the robots' wheels, and 
with $\omega_i$ being its angular velocity.

\subsection{Mixed reality environment}\label{subsec:theplatform}
\begin{figure}
\centering
\subfloat[]{\includegraphics[width=0.24\textwidth]{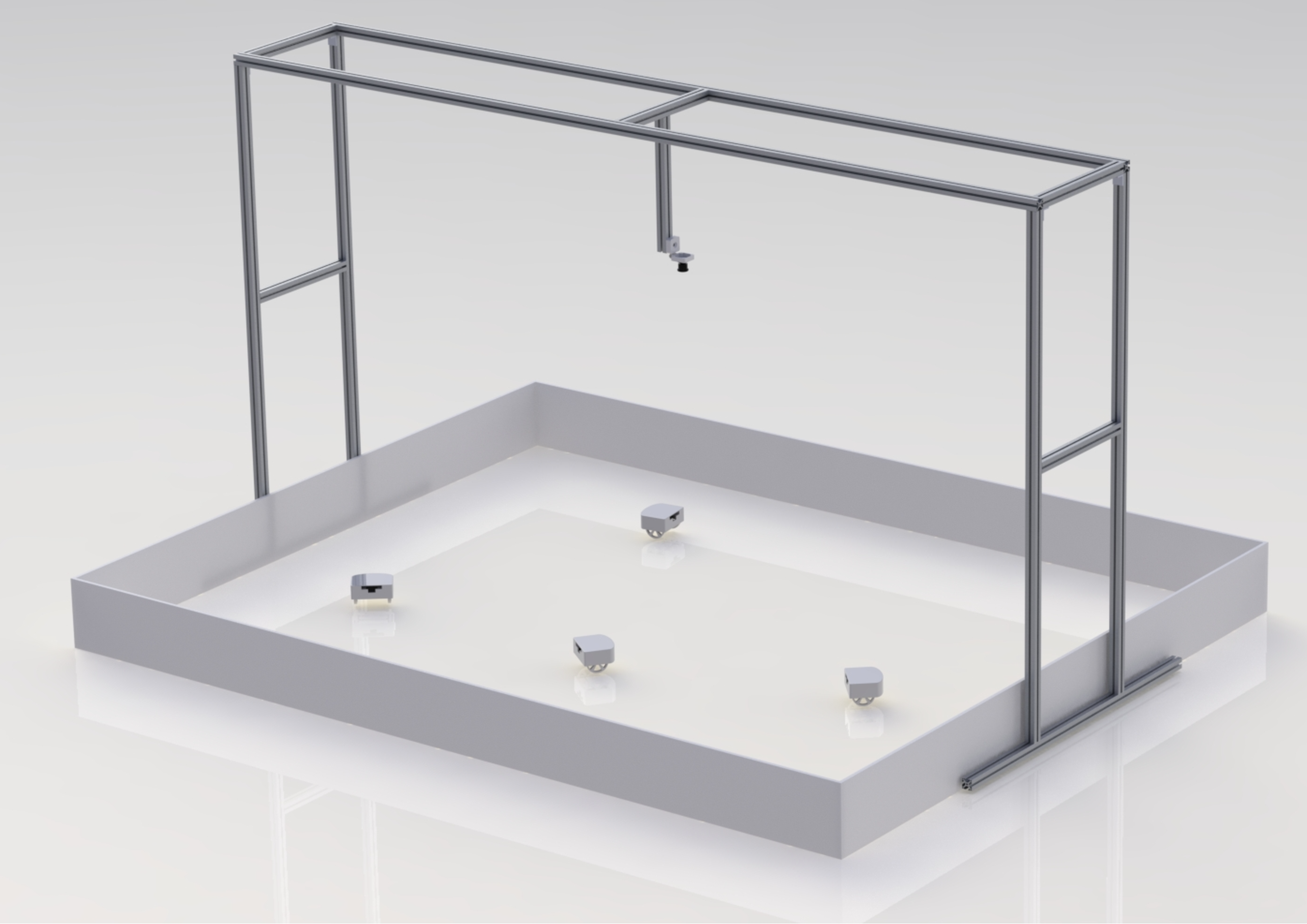}%
\label{subfig::platform}}
\hfil
\subfloat[]{\includegraphics[width=0.24\textwidth]{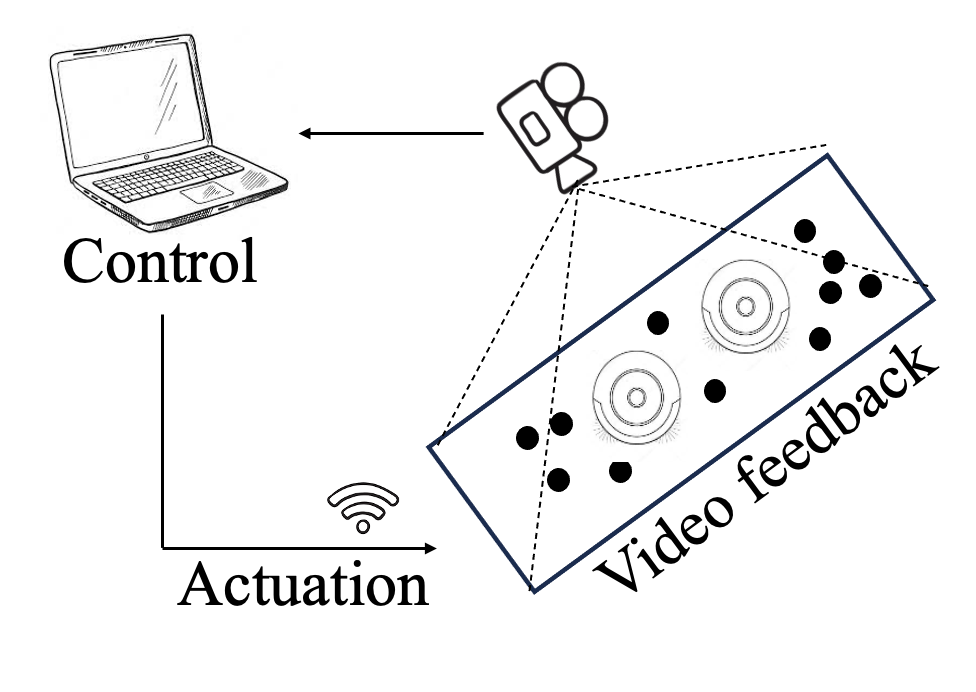}%
\label{subfig::sketch}}
\caption{Experimental platform. (a) A render of the real set-up, with 4 robots moving in the arena, and (b) a sketch of the platform, assuming virtual agents to be the black dots and real robots to be concentric circles.}
\label{fig::theplatform}
\end{figure}
We built the set-up shown in Fig. \ref{subfig::platform}, comprising a set of differential drive robots moving on the ground and an overhead camera (16MP wide-angle camera -- Arducam, placed at 1m height). The camera was placed so that the robots could move in an area of approximately 2 m $\times$ 2 m.
Aruco markers were attached to the robots, so that they could be easily tracked by the camera and perform their pose estimation. A Python program using OpenCV was developed to estimate the robots' pose in each frame. 
The video feed, with all the estimated robots' positions, was given to the central station of the platform, a Dell Aurora ($13^{th}$ Gen Intel® i9 13900KF, 64GB of DDR5 RAM NVIDIA® GeForce RTX™ 4090). Such a machine was also provided with the positions (and eventually velocities) of a user-defined number of virtual agents. In principle, one can choose the specific mathematical model for the virtual agents. Based on the literature about the control of large-scale mobile agents, a reasonable choice is to select their dynamics as that of single or double integrator without any kinematic constraint \cite{bernoff2011primer, viscek1995}.
Using the available information (robots' positions and virtual agents' positions, at least), the central station was in charge of controlling the hybrid swarm of real and virtual agents, according to some user-definable algorithm. Such a control algorithm should be chosen so that the needed information could be estimated by tracking the real robots with a camera, since robots were not equipped with any specific sensor.

The application of any control strategy consists of ($i$) updating the positions of the virtual agents, based on the specific dynamical model that is assumed for them, and ($ii$) computing the control inputs for the real robots and sending them trough a TCP client/server communication protocol on the local Wi-Fi. The idea is sketched in Fig. \ref{subfig::sketch}.
Since collective motion techniques are typically developed for kinematically unconstrained agents, a low-level trajectory tracking control is needed for the robots. 
We used the input/output feedback linearization technique that is proposed in \cite{siciliano} (Chapter 11.6). 

We remark that the set-up we propose is versatile, as it could account for various constraints that can be chosen by the user, like, for instance, limited sensing and obstacles.

\section{High-dimensional Continuification-based Control}\label{sec:theoreticalframework}
We now present the theoretical expansion of our 1D study \cite{maffettone2022continuification} to  higher-dimensional periodic domains. We consider a group of interacting agents, whose dynamics is detailed in Section \ref{subsec:themodel}, within the context of a  density control problem, that is formally given in Section \ref{subsec:probstatement}. Specifically, by appropriately choosing their control inputs, we want
them to displace according to a desired density. The control solution, described in details in Section \ref{sec:control_design}, follows a continuification scheme as the one depicted in Fig. \ref{fig:continuification}. First, we provide some useful notation.

\subsection{Mathematical preliminaries} \label{sec:math_prem}
Here we give some mathematical definitions and concepts that will be used throughout the paper.
We define $\Omega:=[-\pi, \pi]^d$, with $d=1,2,3$ the periodic cube of side $2\pi$. The case $d=1$ coincides with the unit circle, $d=2$ with the periodic square, and $d=3$ with the periodic cube.
We denote by $\Vert h(\cdot, t)\Vert$ the $\mathcal{L}^2$ norm of the function $h: \Omega\times \mathbb{R}_{\geq 0} \to \mathbb{R}$, with respect to its first variable.
For brevity, we will also denote the norm as $\Vert h\Vert$, without explicitly indicating the dependencies. 
We denote with \textquotedblleft{} $*$ \textquotedblright{} the convolution operator. When referring to periodic functions and domains, the convolution needs to be interpreted as its circular version \cite{jeruchim2006simulation}. When one of the functions involved in the convolution is vector valued, the operator is interpreted as the multi-dimensional (circular) convolution.
For PDEs, we denote by $(\cdot)_t$ and $(\cdot)_x$ first order time and space partial derivatives. We use the $\nabla$ operator for vectorial differential operators. Specifically, given a vector valued function $\mathbf{h}$, we denote its gradient as $\nabla \mathbf{h}$, its divergence as $\nabla \cdot \mathbf{h}$, its curl as $\nabla \times \mathbf{h}$, and its Laplacian as $\nabla^2\mathbf{h}$.
We denote by $\mathbb{n} = (n_1,\dots, n_d)$ the $d$-dimensional multi-index, consisting in the tuple of dimension $d$, with $n_i\in\mathbb{Z}$. Thus, $\mathbf{n} = [n_1, \dots, n_d]$ is the row vector associated to $\mathbb{n}$.

\subsection{The model}\label{subsec:themodel}
We consider $N$ dynamical systems moving in $\Omega$. The agents' dynamics is modeled using the \textit{kinematic assumption} \cite{viscek1995, bernoff2011primer} (i.e., neglecting acceleration and considering a drag force proportional to the velocity) and assuming agents are not subject to any non-holonomic constraint. Specifically, we set
\begin{align}\label{eq:d_dimensional_micro}
    \dot{\mathbf{x}}_i = \sum_{k=1}^N \mathbf{f}\left(\{\mathbf{x}_{i}, \mathbf{x}_{k}\}\right) + \mathbf{u}_i, \;\;\;  i=1,\dots,N,
\end{align}
where $\mathbf{x}_i \in \Omega$ is the $i$-th agent's position, and $\{\mathbf{x}_{i}, \mathbf{x}_{k}\}$ is the relative position between agent $i$ and $k$, wrapped to have values in $\Omega$ (see \cite{maffettone2022continuification} for the explicit expression in 1D),
$\mathbf{f}:\Omega\rightarrow \mathbb{R}^d$ is a periodic velocity interaction kernel modeling pairwise interactions between the agents,  and $\mathbf{u}_i$ is a velocity control input designed as to fulfill some control problem. Furthermore, we assume $\mathbf{f}(\mathbf{z}) = -\nabla F(\mathbf{z})$, where $F: \mathbb{R}^d \rightarrow \mathbb{R}$ is a
\textit{soft-core} potential, meaning that $\mathbf{f}(\mathbf{0}) = \mathbf{0}$. The Morse potential, vastly used in the literature \cite{bernoff2011primer, Dorsogna2006}, is a choice of this kind.
We remark that, in the absence of control, agents subject to a repulsive kernel will spread in $\Omega$ until reaching an equilibrium configuration. Agents subject to a Morse-like kernel (long-range attraction and short-range repulsion), will reach an aggregated compact formation (see \cite{bernoff2011primer} for a comprehensive description of the uncontrolled problem with 1D examples). 

Assuming the number of agents $N$ is sufficiently large, we describe the system's collective behavior in terms of the spatio-temporal evolution  of the swarm's density. Hence, we define the density at time $t$ as the scalar function $\rho: \Omega\times \mathbb{R}_{\geq0}\rightarrow \mathbb{R}_{\geq 0}$, such that $\int_{\Omega} \rho(\mathbf{x}, t)\,\mathrm{d}\mathbf{x} = N$ and the integral over a subset of $\Omega$ returns the number of agents in it.

\subsection{Problem statement}\label{subsec:probstatement}
The problem is to select a set of distributed control inputs $\mathbf{u}_i$ acting at the microscopic, agent-level allowing the agents to organize themselves into a desired macroscopic configuration on $\Omega$. 
Specifically, given some desired periodic smooth density profile, $\rho^\text{d}(\mathbf{x}, t)$, associated with the target agents' configuration, the problem can be reformulated as that of finding a set of distributed control inputs $\mathbf{u}_i,\ i=1,2,\dots,N$ in \eqref{eq:d_dimensional_micro} such that
\begin{equation}
    \lim_{t\rightarrow \infty} \Vert{\rho^\text{d}(\cdot, t)}-\rho(\cdot, t)\Vert=0,
\end{equation} 
for agents starting from any initial configuration $\mathbf{x}_i(0)=\mathbf{x}_{i0}, \ i=1,2,\ldots,N$.
This problem is a non-trivial extension to higher dimensions of the one-dimensional problem discussed in \cite{maffettone2022continuification}.

\subsection{Control design} \label{sec:control_design}
We adopt a continuification approach \cite{nikitin2021continuation, maffettone2022continuification, maffettone2023continuification}, consisting in the following steps that are briefly discussed in Sec. \ref{sec:intro}.

$(i)$ Continuification:
in the limit case of an infinite number of agents, we recast the microscopic dynamics of the agents \eqref{eq:d_dimensional_micro} as the mass balance equation \cite{maffettone2022continuification, maffettone2023continuification}
\begin{align}\label{eq:d_dimensional_macro}
    \rho_t(\mathbf{x}, t) + \nabla \cdot \left[\rho(\mathbf{x}, t)\mathbf{V}(\mathbf{x}, t)\right] = q(\mathbf{x}, t), 
\end{align}
where
\begin{align}
    \mathbf{V}(\mathbf{x}, t) = \int_{\Omega} \mathbf{f}\left(\{\mathbf{x}, \mathbf{z}\}\right) \rho(\mathbf{z}, t) \,\mathrm{d}\mathbf{\mathbf{z}} = (\mathbf{f}*\rho)(\mathbf{x}, t).
\end{align}
represents the characteristic velocity field encapsulating the interactions between the agents in the continuum. The scalar function $q$, represents the macroscopic control action. It is written as a mass source/sink for simplifying derivations, but will be in the end recast as an additional velocity field.

For \eqref{eq:d_dimensional_macro} to be well posed, we require the periodicity of $\rho(\mathbf{x}, t)$ on $\partial\Omega$ $\forall t \in \mathbb{R}_{\geq 0}$ and that $\rho(\mathbf{x}, 0) = \rho_0(\mathbf{x})$.
We remark that $\mathbf{V}$ is periodic by construction, as it comes from a circular convolution. Thus, the periodicity of the density is enough to ensure mass is conserved when $q = 0$, i.e., $\left(\int_\Omega \rho(\mathbf{x}, t) \,\mathrm{d}\mathbf{x}\right)_t = 0$ (using the divergence theorem and exploiting the periodicity of the flux). 

\begin{rem}
    We do not assume the agents' dynamics to be linear and interactions to take place on a spatially-invariant lattice as done in \cite{nikitin2021continuation}, where some useful heuristics extensions to nonlinear systems and different topologies are presented.
\end{rem}

$(ii)$ Macroscopic control design: 
we assume the desired density profile obeys to the mass conservation law
\begin{align}\label{eq:d_dimensional_ref_dyn}
    \rho^\mathrm{d}_t(\mathbf{x}, t) + \nabla \cdot \left[\rho^\mathrm{d}(\mathbf{x}, t)\mathbf{V}^\mathrm{d}(\mathbf{x}, t)\right] = 0,
\end{align}
where 
\begin{align}
    \mathbf{V}^\mathrm{d}(\mathbf{x}, t) = \int_{\Omega} \mathbf{f}\left(\{\mathbf{x, \mathbf{z}}\}\right) \rho^\mathrm{d}(\mathbf{z}, t) \,\mathrm{d}\mathbf{\mathbf{z}} = (\mathbf{f}*\rho^\mathrm{d})(\mathbf{x}, t).
\end{align}
Periodic boundary conditions and initial condition for \eqref{eq:d_dimensional_ref_dyn} are set similarly to those of \eqref{eq:d_dimensional_macro}.
Furthermore, we define the error function $e(\mathbf{x}, t) := \rho^\mathrm{d}(\mathbf{x}, t)-\rho(\mathbf{x}, t)$.

\begin{thm}[Macroscopic convergence]\label{th_GAS_d_dim}
    Choosing 
    \begin{multline}\label{eq:d_dimensional_control}
        q(\mathbf{x}, t) = K_\mathrm{p} e(\mathbf{x}, t) - \nabla \cdot \left[e(\mathbf{x}, t) \mathbf{V}^\mathrm{d}(\mathbf{x}, t) \right] \\- \nabla \cdot \left[\rho(\mathbf{x}, t) \mathbf{V}^\mathrm{e}(\mathbf{x}, t) \right],
    \end{multline}
    where $K_\mathrm{p}$ is a positive control gain and $\mathbf{V}^\mathrm{e}(\mathbf{x}, t) = (\mathbf{f}*e)(\mathbf{x}, t) $, the error dynamics globally asymptotically converges to 0 
    \begin{align}\label{eq:d_dim_lim}
        \lim_{t\to\infty} e(\mathbf{x}, t) = 0 \;\;\; \forall \,e(\mathbf{x}, 0).
    \end{align}
\end{thm}
\begin{proof}
    We can compute the error dynamics by subtracting \eqref{eq:d_dimensional_macro} from \eqref{eq:d_dimensional_ref_dyn}, resulting in
    \begin{multline}\label{eq:err_dyn_1}
        e_t(\mathbf{x}, t) + \nabla \cdot \left[\rho^\mathrm{d}(\mathbf{x}, t) \mathbf{V}^\mathrm{d}(\mathbf{x}, t) \right] - \\\nabla \cdot \left[\rho(\mathbf{x}, t) \mathbf{V}(\mathbf{x}, t) \right] = -q(\mathbf{x}, t).
    \end{multline}
    The error function $e(\mathbf{x}, t)$ is periodic on $\partial \Omega$ $\forall t\in\mathbb{R}_{\geq0}$ and $e(\mathbf{x}, 0) = \rho^\mathrm{d}(\mathbf{x}, 0)-\rho(\mathbf{x}, 0)$. Then, taking into account that $\rho = \rho^\mathrm{d}-e$, and $\mathbf{V} = \mathbf{V}^\mathrm{d}-\mathbf{V}^\mathrm{e}$, we rewrite \eqref{eq:err_dyn_1} as
    \begin{multline}
        e_t(\mathbf{x}, t) + \nabla \cdot \left[e(\mathbf{x}, t) \mathbf{V}^\mathrm{d}(\mathbf{x}, t) \right] + \\\nabla \cdot \left[\rho(\mathbf{x}, t) \mathbf{V}^\mathrm{e}(\mathbf{x}, t) \right]= -q(\mathbf{x}, t).
    \end{multline}
    Plugging in \eqref{eq:d_dimensional_control}, we get
    \begin{align}
        e_t(\mathbf{x}, t) = -K_\mathrm{p}e(\mathbf{x}, t).
    \end{align}
    Since  $K_\mathrm{p}>0$, \eqref{eq:d_dim_lim} holds.
\end{proof}

$(iii)$ Discretization and microscopic control:
\label{subsec:discretization}
in order to dicretize the macroscopic control action $q$, we first recast the macroscopic controlled model as
\begin{align}\label{eq:controlled_by_U}
    \rho_t(\mathbf{x}, t) + \nabla \cdot \left[\rho(\mathbf{x}, t) \left(\mathbf{V}(\mathbf{x}, t) + \mathbf{U}(\mathbf{x}, t)\right)\right] = 0,
\end{align}
where $\mathbf{U}$ is a controlled velocity field, in which we want to incorporate the control action.
Equation \eqref{eq:controlled_by_U} is equivalent to \eqref{eq:d_dimensional_macro}, if
\begin{align}\label{eq:divergence_of_U}
    \nabla \cdot \left[\rho(\mathbf{x}, t)\mathbf{U}(\mathbf{x}, t)\right] = -q(\mathbf{x}, t).
\end{align}
In contrast to the case where $d = 1$ discussed in reference \cite{maffettone2022continuification}, equation \eqref{eq:divergence_of_U} is insufficient to uniquely determine $\mathbf{U}$ from $q$ since it represents only a scalar relationship (as the divergence returns a scalar function). Hence, we define the flux $\mathbf{w}(\mathbf{x}, t):= \rho(\mathbf{x}, t)\mathbf{U}(\mathbf{x}, t)$, and close the problem by adding an extra differential constraint on the curl of $\mathbf{w}$. Namely, we consider the set of equations
\begin{align}\label{eq:w}
    \begin{cases}
        \nabla \cdot \mathbf{w}(\mathbf{x}, t) = -q(\mathbf{x}, t)\\
    \nabla \times \mathbf{w}(\mathbf{x}, t) = 0
    \end{cases}
\end{align}
For problem \eqref{eq:w} to be well posed, we require $\mathbf{w}(\mathbf{x}, t)$ to be periodic on $\partial \Omega$. 
Notice that \eqref{eq:w} is a purely spatial problem, as no time derivatives are involved.
We also remark that the choice of closing the problem using the irrotationality condition is arbitrary, and other closures can be considered. This specific one allows not to introduce vorticity into the velocity field we are looking for. Since $\Omega$ is simply connected, and $\nabla \times\mathbf{w} = 0$, we can express $\mathbf{w}$ using the scalar potential $\varphi$. Specifically, we pose $\mathbf{w} (\mathbf{x}, t) = -\nabla\varphi(\mathbf{x}, t)$, making the zero-curl condition always fulfilled. Then, substituting this into the divergence relation in \eqref{eq:w}, we can recast \eqref{eq:w} as the Poisson equation
\begin{align}\label{eq:Poisson}
    \nabla^2 \varphi (\mathbf{x}, t) = q (\mathbf{x}, t).
\end{align}

Problem \eqref{eq:Poisson} is characterized by the periodicity of $\nabla \varphi(\mathbf{x}, t)$ on $\partial \Omega$. We wish to remark the analogy with the derivation of the Poisson equation in the context of the electrostatic field \cite{griffiths2005introduction}. Equation \eqref{eq:Poisson}, together with its boundary conditions, defines $\varphi$ up to a constant $C$. Since we are interested in computing $\mathbf{w} = -\nabla \varphi$, the value of $C$ is irrelevant.
We solve the Poisson problem \eqref{eq:Poisson} in $\Omega$ using the Fourier series. Then, writing the Fourier series of $\varphi$ in $\Omega$, we get
\begin{align}\label{eq:phi}
    \varphi(\mathbf{x}) = \sum_{\mathbb{m}\in \mathbb{Z}^d} \gamma_\mathbb{m} \,\mathrm{e}^{j\mathbf{m}\cdot \mathbf{x}} + C,
\end{align}
where $\mathbb{m}$ is a multi-index, $\mathbf{m}$ is the row vector associated to this multi-index, $\gamma_\mathbf{m}$ is the $\mathbb{m}$-th Fourier coefficient, $j$ is the imaginary unit, and $\mathbf{x}$ is assumed to be a column. Given this expression for the potential, we write its Laplacian as
\begin{align}\label{eq:fourier_laplacian}
    \nabla^2\varphi(\mathbf{x}) = \sum_{\mathbb{m}\in \mathbb{Z}^d} \gamma_\mathbb{m} \Vert \mathbf{m}\Vert^2 \mathrm{e}^{j\mathbf{m}\cdot \mathbf{x}}.
\end{align}
Next, we can  apply Fourier series to the known function $q$, resulting in
\begin{align}\label{eq:fourier_q}
    q(\mathbf{x}) = \sum_{\mathbb{m}\in \mathbb{Z}^d} c_\mathbb{m} \,\mathrm{e}^{j\mathbf{m}\cdot \mathbf{x}}, 
\end{align}
where, since at time $t$ the function $q$ is known, we can also express the coefficients as
\begin{align}
    c_\mathbb{m} = \frac{1}{(2\pi)^d}\int_\Omega q(\mathbf{x}) \mathrm{e}^{-j\mathbf{m}\cdot \mathbf{x}}\,\mathrm{d}\mathbf{x}.
\end{align}
Then, recalling \eqref{eq:Poisson}, we can express the coefficients of the Fourier series of the potential $\varphi$ as
\begin{align}
    \gamma_\mathbb{m} = -\frac{c_\mathbb{m}}{\Vert \mathbf{m}\Vert^2}.
\end{align}

Hence $\mathbf{w} = -\nabla \varphi$ and, consequently, $\mathbf{U} = \mathbf{w}/\rho$. Such derivations need to take place at each $t$. From the implementation viewpoint, when computing $\varphi$, and consequently $\mathbf{w}$, we approximate it only considering the first $M$ (with $M$ sufficiently large) terms of the infinite summations in \eqref{eq:phi}. 

Then, we can compute the microscopic control inputs for the discrete set of agents by spatially sampling  $\mathbf{U}(\mathbf{x}, t)$, that is
\begin{align}
    \mathbf{u}_i(t) = \mathbf{U}(\mathbf{x}_i, t), \;\;\;\; i=1,2,\dots,N.
\end{align}
Notice that our discretization procedure is different from the one that is proposed in \cite{nikitin2021continuation}.
\begin{rem}
    The macroscopic control action $q$ is based on non-local terms like $\mathbf{V}^\mathrm{d}$ and $\mathbf{V}^\mathrm{e}$, making the control action exerted at $\mathbf{x}$ depending on the error everywhere else in $\Omega$. The input $\mathbf{u}_i$ can be approximated in terms of local information, since the assumption of unlimited sensing is practically mitigated by assuming a vanishing interaction kernel. We refer to \cite{maffettone2023continuification}, \cite{maffettone2024high} for analytical results about limited sensing.
\end{rem}
\begin{rem}
    The macroscopic velocity field $\mathbf{U}$ is well-defined only when $\rho \neq 0$. This is indeed a fair assumption, as finally we will estimate the density by the agents position with an estimation kernel of our choice. Moreover, as $\mathbf{U}$ will be sampled at the agents locations, i.e. where the density is different from 0, we know $\mathbf{U}$ is well defined where it is needed.
\end{rem}
\begin{rem}
    The proposed technique differs from its one-dimensional counterpart in \cite{maffettone2022continuification} for the steps following \eqref{eq:divergence_of_U}. In particular, if $d=1$, \eqref{eq:divergence_of_U} can be spatially integrated to uniquely determine $U$ (one-dimensional version of $\mathbf{U}$) from $q$.
\end{rem}
\section{Validation of the control approach via the new experimental platform}\label{sec:demonstration}
\begin{figure}
    \centering
    \includegraphics[width=0.45\textwidth]{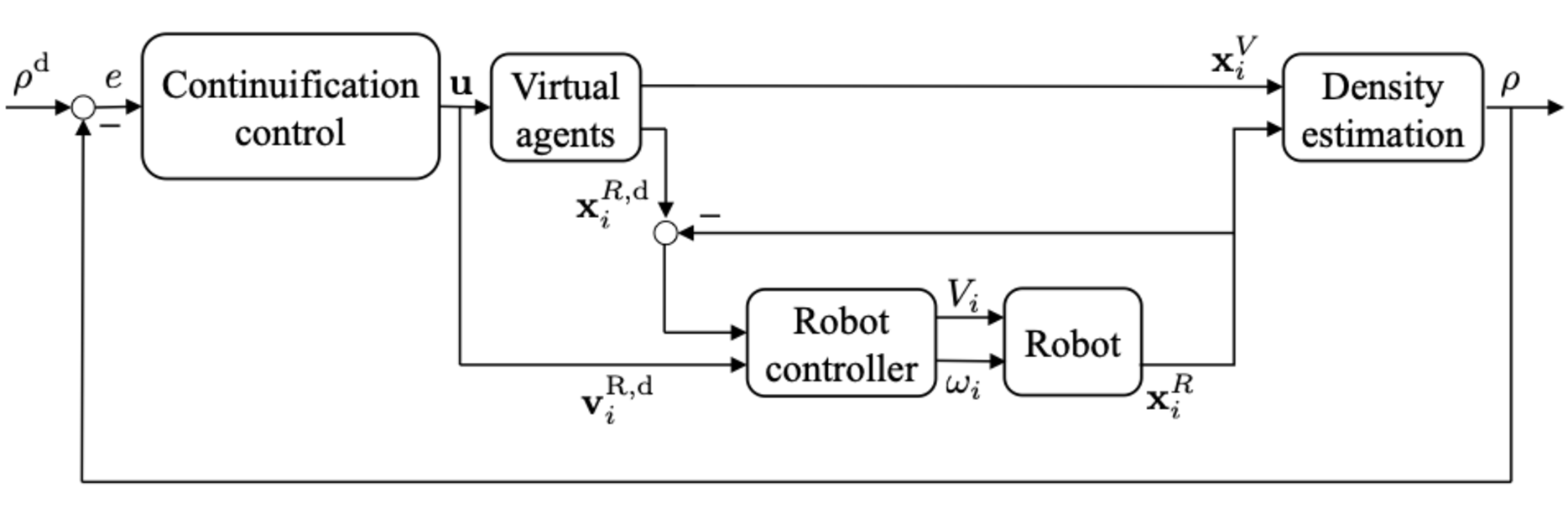}
    \caption{Control scheme for robot $i$. By measuring the overall density of the swarm, the continuification control inputs can be used to give the robots a desired position and velocity to track.}
    \label{fig:multi_rob_cont}
\end{figure}
Next, we experimentally validate the higher-dimensional continuification control strategy proposed in Section \ref{sec:control_design} to steer the collective behavior of a swarm of robots in the plane. In so doing, we also demonstrate the use of our experimental platform for evaluating the performance of control algorithms. To this aim,  we fix $d=2$, making $\Omega$ the periodic square. For modeling pairwise interactions between the agents, we choose a periodic soft-core repulsive kernel, based on its non-periodic version
\begin{align}\label{eq:kernel}
    \mathbf{f}(\mathbf{x}) = \begin{cases}
        \frac{\mathbf{x}}{\Vert \mathbf{x}\Vert} \mathrm{e}^{-\frac{\Vert \mathbf{x}\Vert}{L}} \;\;&\mathrm{if}\;\Vert \mathbf{x}\Vert\neq 0\\
        0 &\mathrm{otherwise}.
    \end{cases}
\end{align}
\noindent The periodization of the kernel consists in an infinite series extending the non-periodic kernel in every direction \cite{stein2011fourier}. Since no closed form was found, we approximate it by truncating the series. Moreover, we fix $L=1$.

In what follows, we always refer to a Cartesian coordinate system, like the one considered for the individual kinematics. For each experimental trial we consider that agents start on a perfect square lattice, meaning that the initial density is constant and, in particular, $\rho(\mathbf{x}, 0) = N/(2\pi)^2$.
As for the desired density to achieve, we choose the 2D Von-Mises function
\begin{align}\label{eq:2Dvon_mises}
    \rho^\mathrm{d}(\mathbf{x}) = Z \,\mathrm{exp}\{\mathbf{k}^T\, \mathbf{c}_1(\mathbf{x}, \mu, \nu) + c_2(\mathbf{x}, \mu, \mu) \,\mathbf{I}_2 \, c_2^T(\mathbf{x}, \nu, \nu)\}
\end{align}
where $\mathbf{k} = [k_1, k_2]^T$ is the vector of the concentration coefficients, $\mu$ and $\nu$ are the means along the two directions, 
$\mathbf{c_1}(\mathbf{x}, a, b) = [\cos(x_1 - a), \cos(x_2 - b)]$ and $\mathbf{c_2}(\mathbf{x}, a, b) = [\cos(x_1 - a), \sin(x_2 - b)]$ (with $a, b \in \Omega$), where
$x_1$ and $x_2$ are the components of $\mathbf{x}$ in the Cartesian coordinate system, and $\mathbf{I}_2$ is the second order identity matrix. $Z$ is a normalization coefficient, to allow $\rho^\mathrm{d}$ to sum to the total number of agents $N$.
To assess the performance in different scenarios, we also take into account the case where the desired density is multimodal, that is the combination of several densities like \eqref{eq:2Dvon_mises}. To address tracking scenarios as well, we study the case where the means, $\mu$ and $\nu$, in \eqref{eq:2Dvon_mises} are time varying. 
We remark that the scenarios we consider mimic more classical microscopic problems of spatial organization. For instance, density regulation to Gaussian-like profiles can be seen as rendez-vouz problems \cite{ando1999distributed}, while tracking cases as formation control ones \cite{oh2015survey}.

The overall control scheme for the hybrid swarm is shown in Fig. \ref{fig:multi_rob_cont}. Specifically, while virtual agents' positions can be updated purely using the technique described in Section \ref{sec:theoreticalframework}, for the differential drive robots, that are kinematically constrained (see Section \ref{subsec:therobots}), such a method needs to be integrated with an ad-hoc controller for tracking problems. As previously mentioned, we used the input/output feedback linearization technique in \cite{siciliano} (Chapter 11.6). The integration is performed by using the continuification method to compute the desired position and velocity of the robot, that is then tracked with its embedded controller. In the case where only real robots are present, the blocks regarding virtual agents in the scheme in Fig. \ref{fig:multi_rob_cont} shall be omitted.
\begin{figure}
\centering
\subfloat[]{\includegraphics[width=0.24\textwidth]{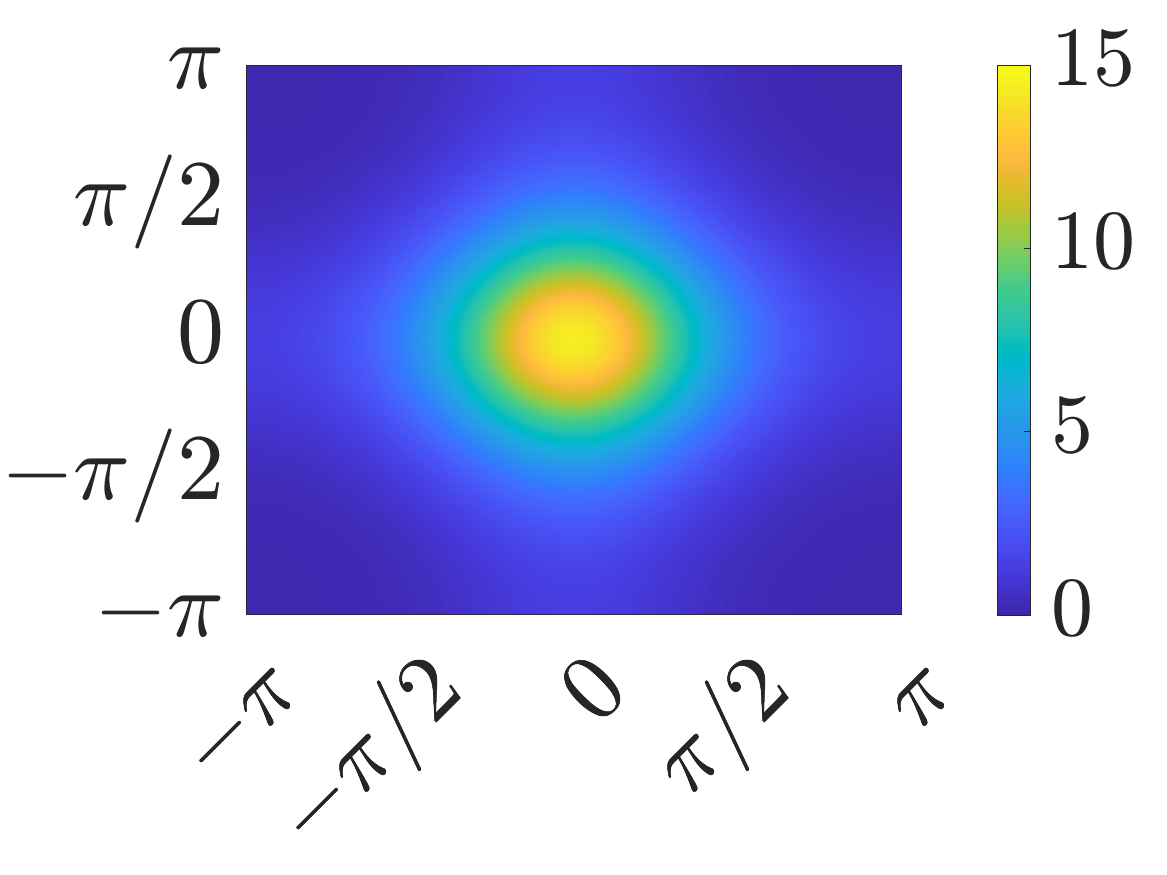}%
\label{subfig:monomod_reg_des_den}}
\hspace{0.2cm}
\subfloat[]{\includegraphics[width=0.18\textwidth]{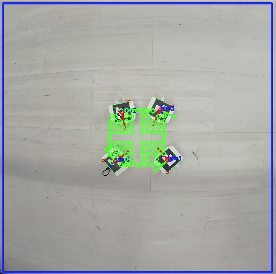}%
\label{subfig:monomod_reg_fin_disp}}

\subfloat[]{\includegraphics[width=0.24\textwidth]{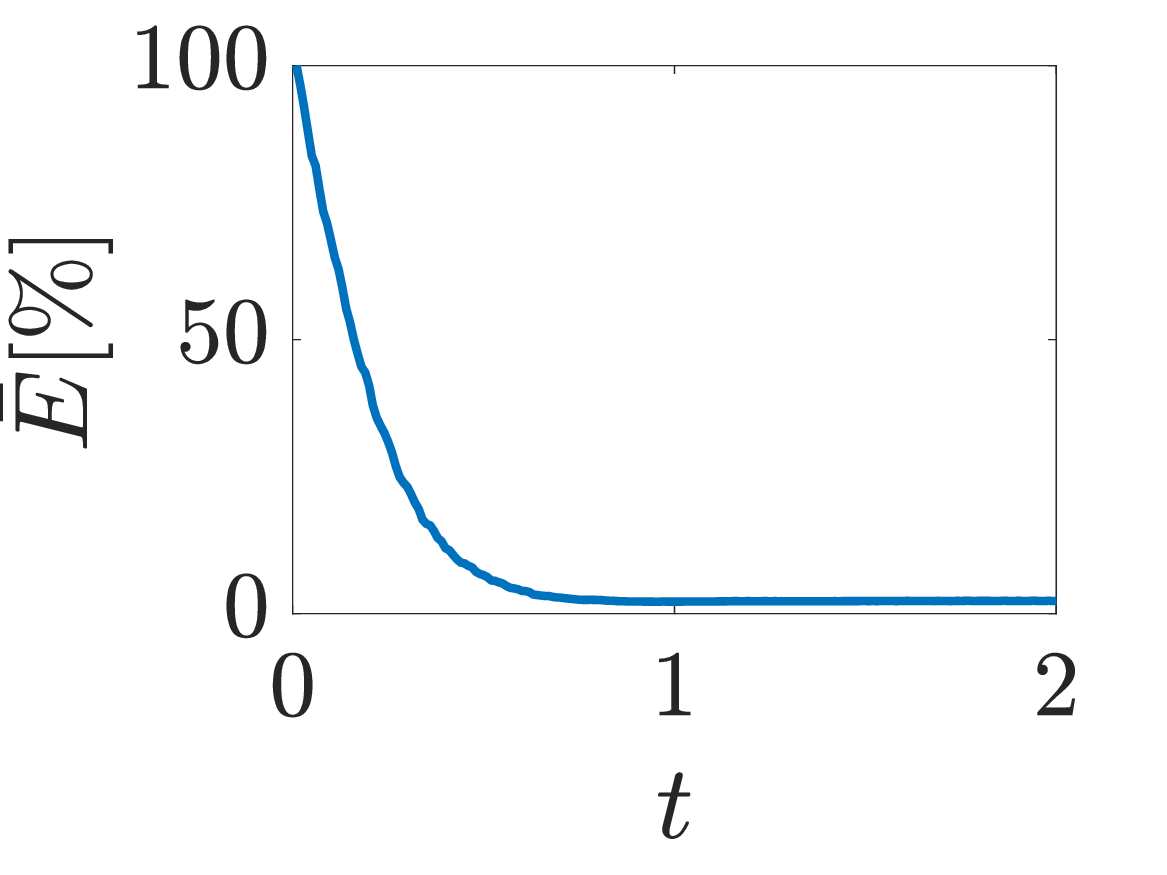}
\label{subfig:monomod_reg_dkl}}
\subfloat[]{\includegraphics[width=0.24\textwidth]{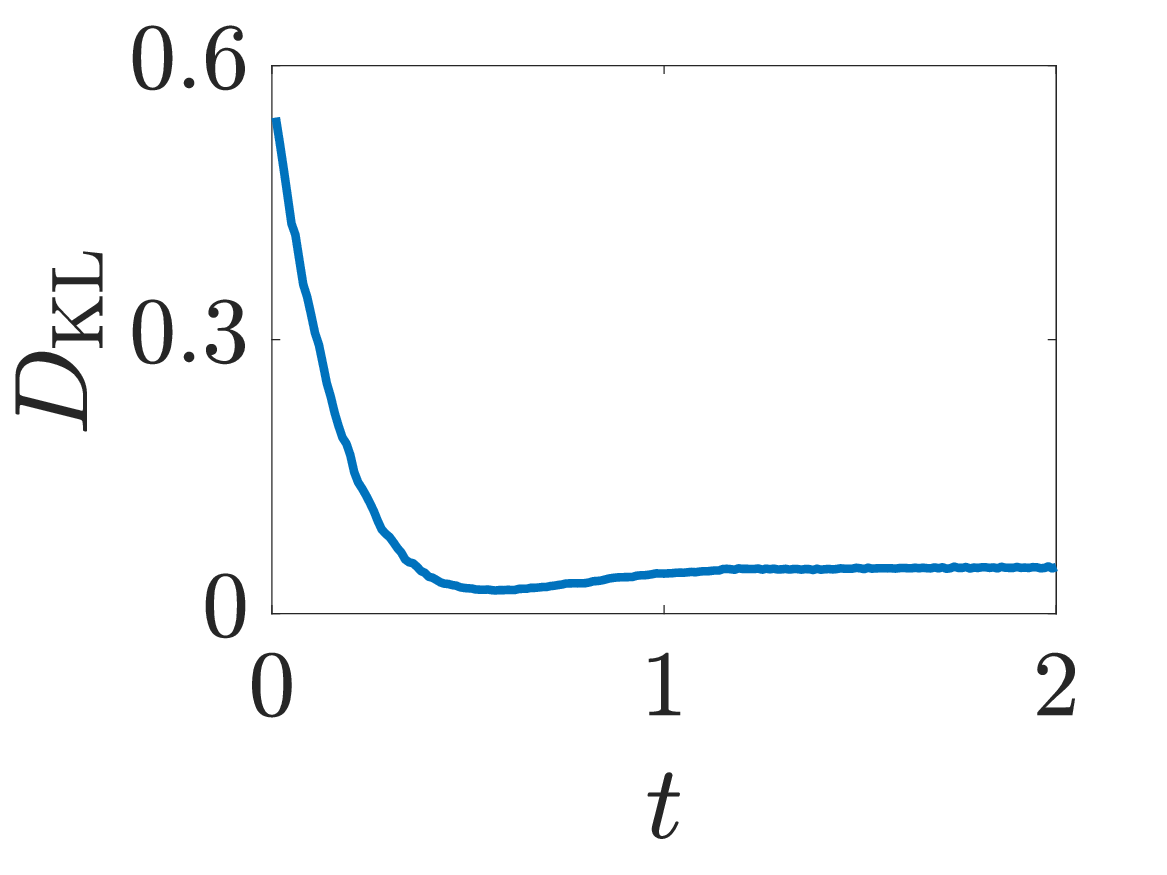}}
\label{subfig:monomod_reg_dkl2}
\caption{Monomodal regulation. (a) Desired density, (b) steady-state configuration of the swarm, (c) percentage error in time, (d) KL divergence in time.}
\label{fig:monomodal_regulation}
\end{figure}
To adapt the assumption on the periodicity of the domain to the experiments and avoid real robots to try to cross the domain's boundaries, we defined a fictitious periodic extended domain (double sized with respect to the effective arena where robots move). The arena where agents move is the inner part of such an extended domain. To avoid agents to go out of the arena (that is the inner part of the domain), the desired density is set as the actual one in the inner part of the domain, and is then extended to be almost zero elsewhere (i.e., in the arena fictitious extension). 

We characterize the experiments recording $\Vert e \Vert^2$ in time. Specifically, the performance of each trial is assessed in terms of the percentage error
\begin{align}
    \bar{E}(t) =\frac{\Vert e(\cdot, t)\Vert^2}{\max_t\Vert e(\cdot, t)\Vert^2}100.
\end{align}
The value of $\bar{E}$ at the end of the trial is the remaining percentage $\mathcal{L}^2$ error. 
Trials are also characterized using the Kullback-Leibler (KL) divergence, as often done for density control problems \cite{gagliardi2022probabilistic}.

For each trial, we considered a sample of $N = 100$ agents (96 virtual agents and 4 real robots), and we discretized \eqref{eq:d_dimensional_micro} (modeling the motion of the virtual agents and the desired positions for the robots) using forward Euler with a non-dimensional time step $\Delta t = 0.01$. This corresponds to the camera frame rate of  20 frames per second (FPS) in the experiments, at which the control algorithm is running. Thus, the unit non-dimensional time in any of our graphs corresponds to 5 s. The spatial domain is discretized into a regular mesh of $200\times200$ cells. We remark that virtual agents are indeed not constrained to move on such a mesh, and that it is only used for defining functions such as the desired and effective density of the swarm. We also remark that spatial measures are adapted to consider that the region where robots are moving coincides with the definition of $\Omega$.

\begin{figure}
\centering
\subfloat[]{\includegraphics[width=0.24\textwidth]{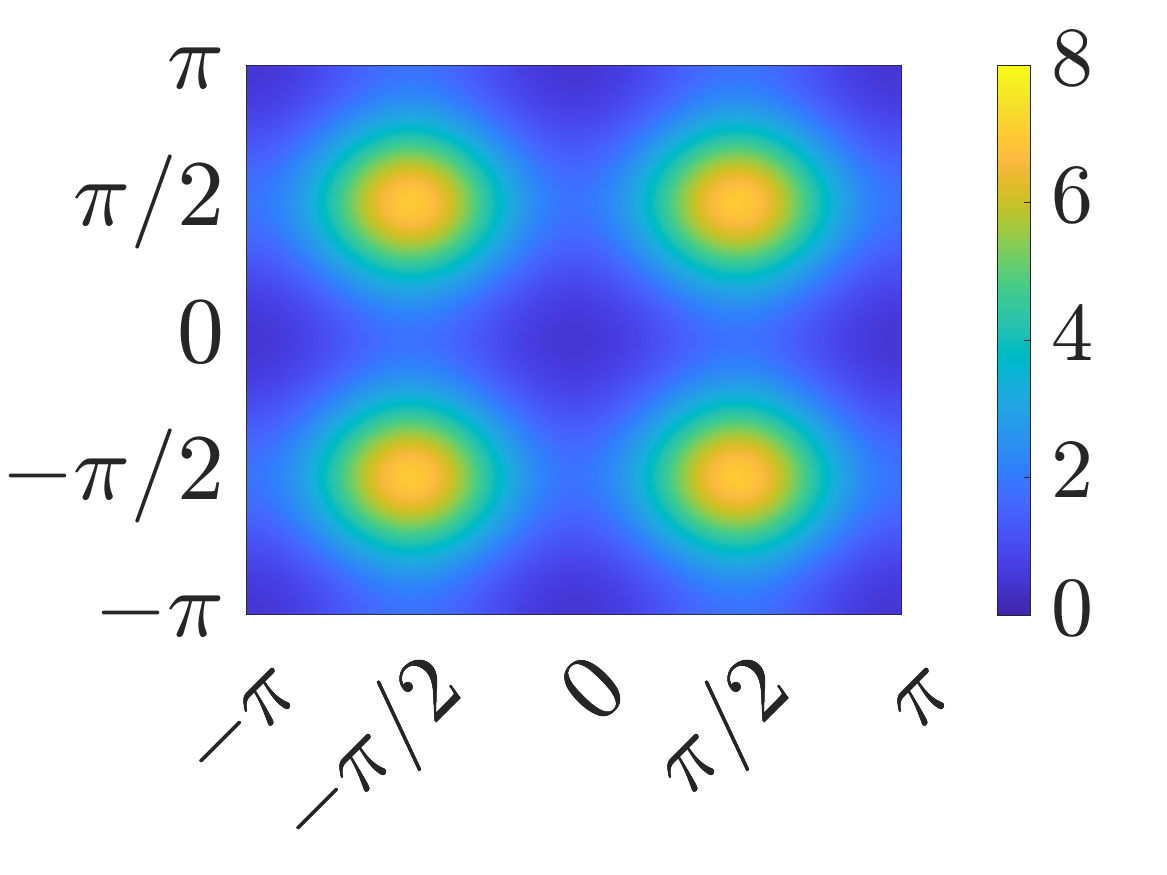}%
\label{subfig:multimod_reg_des_den}}
\hspace{0.2cm}
\subfloat[]{\includegraphics[width=0.18\textwidth]{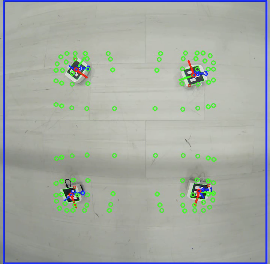}%
\label{subfig:multimod_reg_fin_disp}}

\subfloat[]{\includegraphics[width=0.24\textwidth]{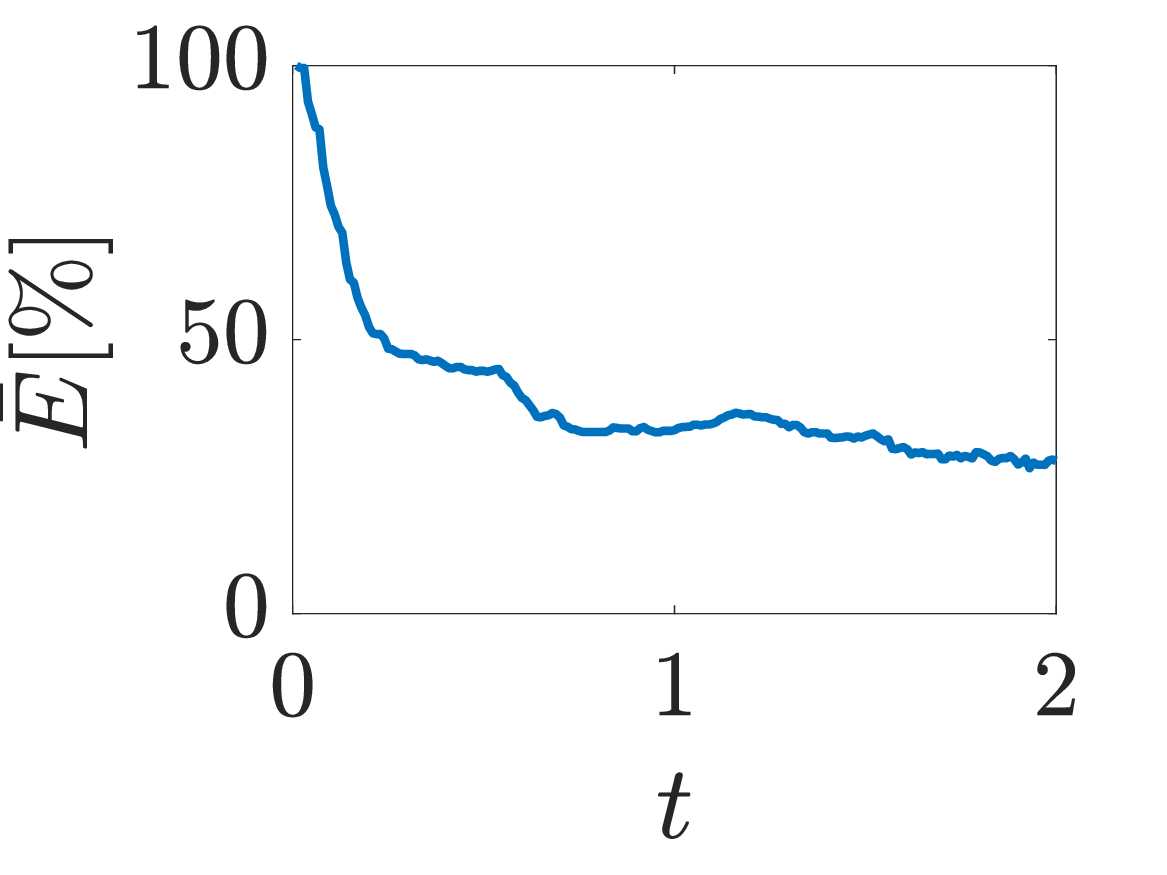}%
\label{subfig:multimod_reg_dkl}}
\subfloat[]{\includegraphics[width=0.24\textwidth]{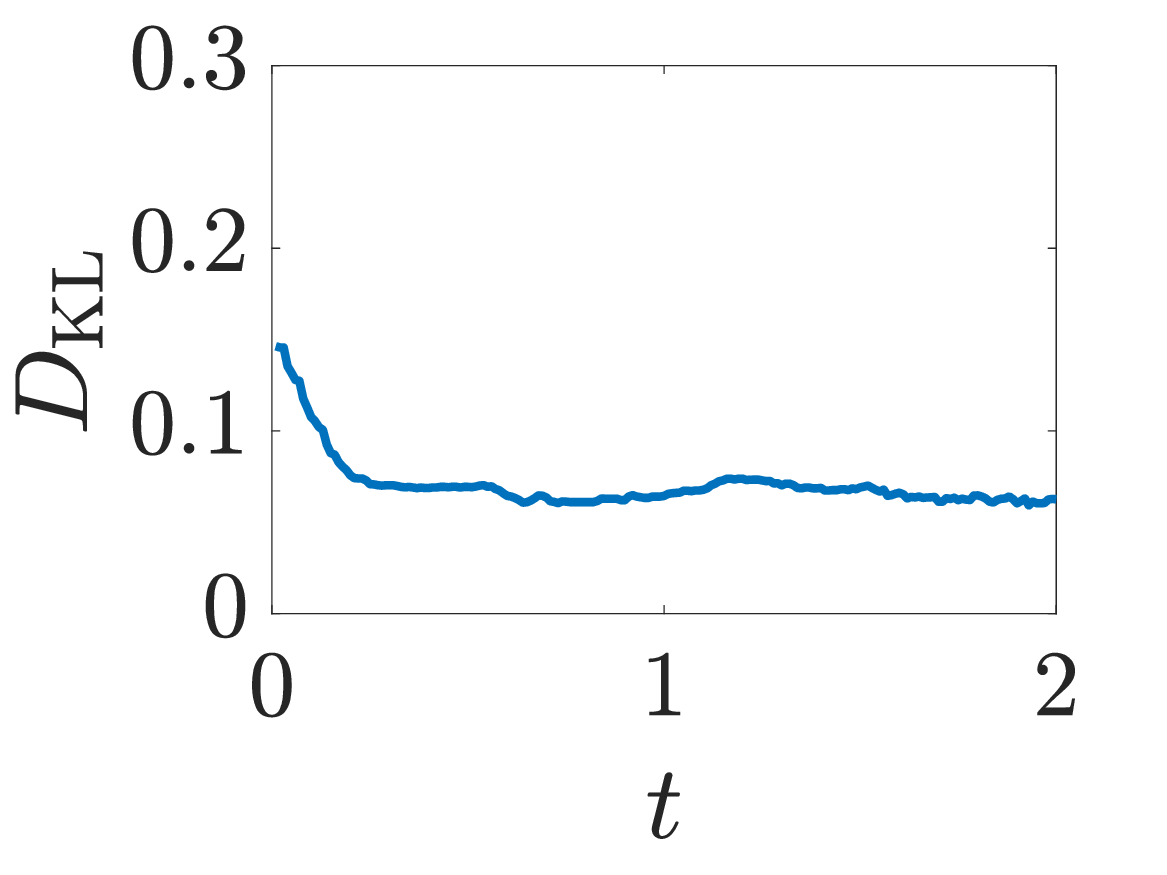}}%
\label{subfig:multimod_reg_dkl2}
\caption{Multimodal regulation. (a) Desired density, (b) steady-state configuration of the swarm, (c) percentage error in time, (d) KL divergence in time.}
\label{fig:multimodal_regulation}
\end{figure}
\subsection{Experimental trials}
Here, we detail our experiments, whose videos are available on Github at \href{https://github.com/Dynamical-Systems-Laboratory/ContinuificationControl}{https://github.com/Dynamical-Systems-Laboratory/ContinuificationControl.git}.
\paragraph{Monomodal regulation}
We want the hybrid swarm to start from an initial constant density and aggregate towards the von Mises function that is depicted in Fig. \ref{subfig:monomod_reg_des_den}, which is characterized by $\mu=\nu=0$, and $\kappa_1=\kappa_2 = 1.5$ (see \eqref{eq:2Dvon_mises}). Such a desired configuration consists in a clustered formation about the origin of $\Omega$. The final formation that is achieved by the swarm is reported in Fig. \ref{subfig:monomod_reg_fin_disp}, while the time evolution of $\bar{E}$ is shown in Fig. \ref{subfig:monomod_reg_dkl}. We record a steady-state value of $\bar{E}$, that is the residual percentage $\mathcal{L}^2$ error, of approximately 2\%. In Fig. \ref{subfig:monomod_reg_dkl2}, we report the time evolution of the KL divergece.

\paragraph{Multimodal regulation}
We consider the swarm to start from an initial constant density and aggregate towards the combination of four von Mises functions as the one in \eqref{eq:2Dvon_mises} (see Fig. \ref{subfig:multimod_reg_des_den} for a graphical representation). The concentration coefficients of all the modes is set to 2, and the mean values are  $\mu_1 = \mu_2= -\pi/2$, $\mu_3=\mu_4=\pi/2$,  $\nu_1 = \nu_2=\pi/2$, and  $\nu_3 = \nu_4=\pi/2$. This desired density consists of four clusters of agents symmetrically displaced around the origin. The final formation is reported in Fig. \ref{subfig:multimod_reg_fin_disp}, while the time evolution of $\bar{E}$ is shown in Fig. \ref{subfig:multimod_reg_dkl}. The final value of $\bar{E}$ is below 30\%. In Fig. \ref{subfig:multimod_reg_dkl2} we show the time evolution of the KL divergence.

\paragraph{Monomodal tracking}
\begin{figure*}
\centering
\subfloat[]{\includegraphics[width=0.24\textwidth]{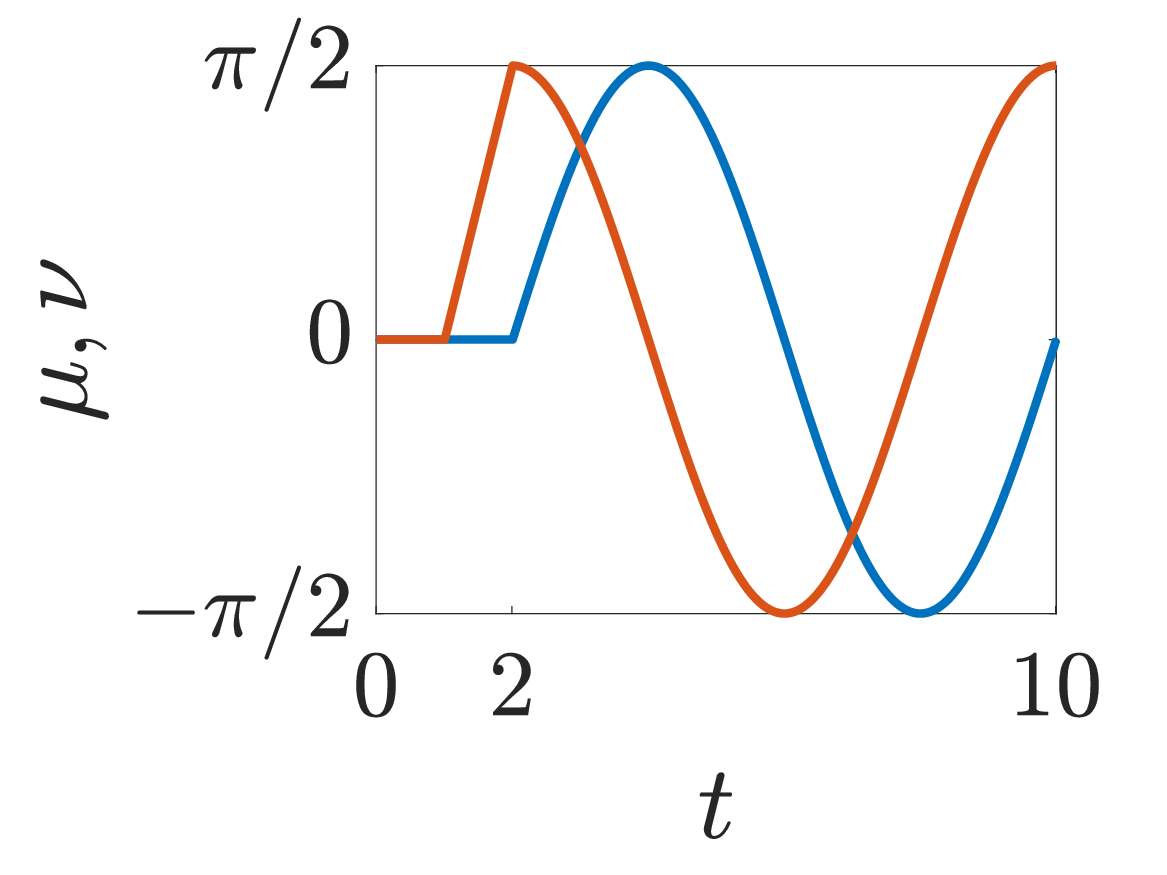}%
\label{subfig:munu_mono_track}}
\subfloat[]{\includegraphics[width=0.24\textwidth]{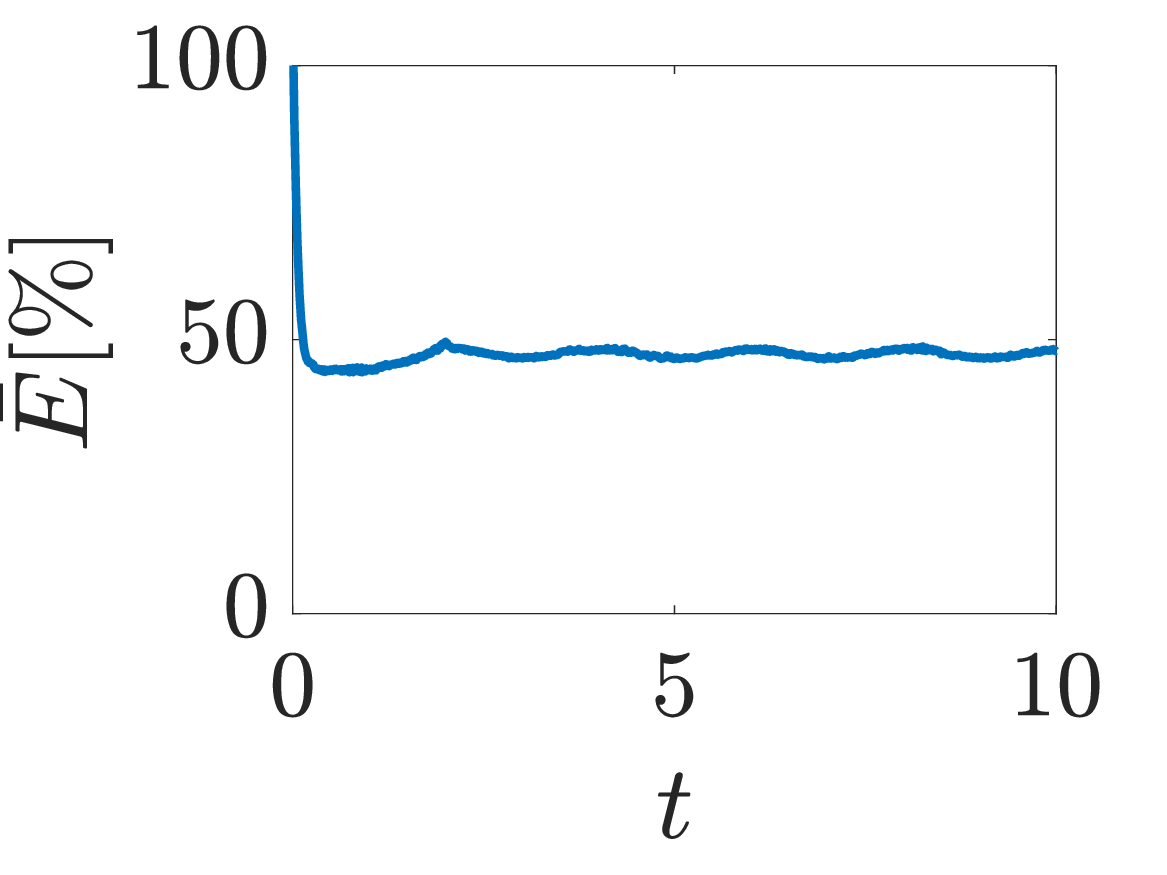}%
\label{subfig:dkl_mono_trakc}}
\subfloat[]{\includegraphics[width=0.24\textwidth]{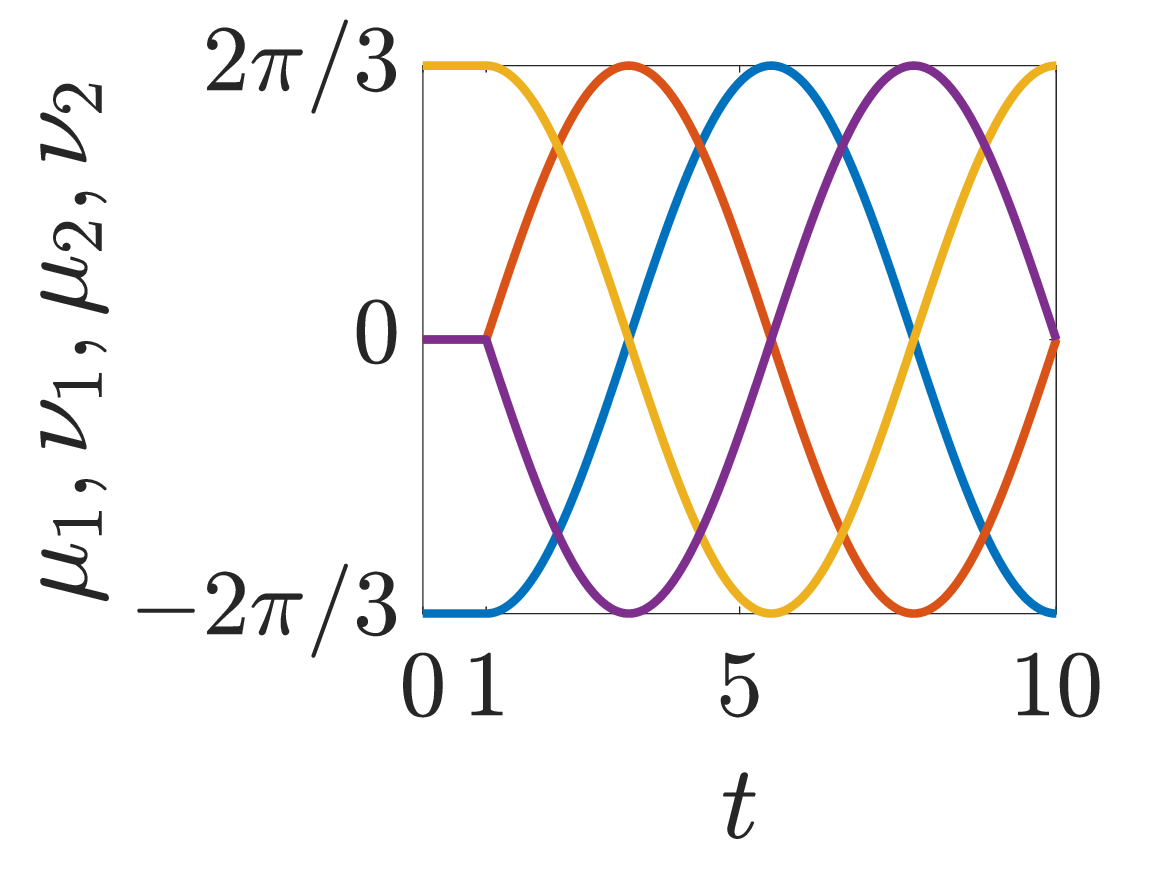}%
\label{subfig:mu1nu1_multi_track}}
\subfloat[]{\includegraphics[width=0.24\textwidth]{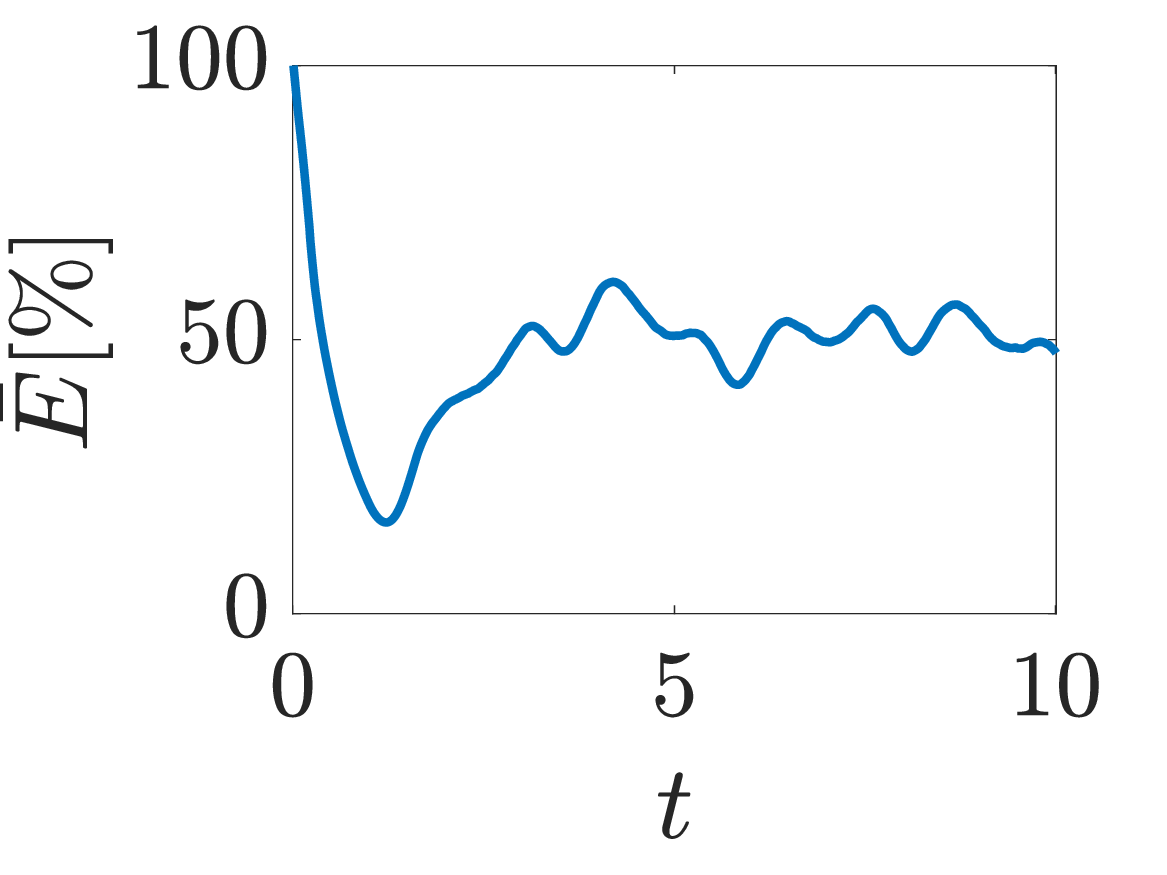}%
\label{subfig:dkl_multi_track}}
\caption{Tracking experiments. (a) Time evolution of the means of the monomodal time variant desired density to track, (b) percentage error in time during the monomodal tracking trial, (c) time evolution of the means of the two modes of the desired von Mises functions in the multimodal tracking trial (fist mode blue and orange, second mode yellow and purple), and (d) percentage error in time during the multimodal tracking trial.}
\label{fig:tracking}
\end{figure*}
Here, we focus on a monomodal tracking scenario, where the desired density is a 2D von Mises function, whose means are time varying, see \eqref{eq:2Dvon_mises}. Specifically, we consider $\mu(t)$ and $\nu(t)$ behaving as in Fig. \ref{subfig:munu_mono_track}, while the concentration coefficients are kept constant and equal to 1. Such a desired density is centered at the origin for $t\leq1$. Then, it starts moving at constant velocity towards a side of the domain and then on the circle of radius $\pi/2$. We report the results of the trial in Fig. \ref{subfig:dkl_mono_trakc}, where the evolution of $\bar{E}$ is shown. Specifically, its steady-state value is below 50\%. For brevity, we do not report the KL divergence in time, which remains below 0.25.

\paragraph{Multimodal tracking} 
Here, we consider a multimodal tracking case, where two von Mises functions with constant concentration coefficients of 2.2 orbitate on the circle of radius $2\pi/3$, after remaining still at two sides of the domain for $t\leq 1$. Specifically, $\mu_1(t)$, $\nu_1(t)$ and $\mu_2(t)$, $\nu_2(t)$, the means of the two von Mises functions, evolve as in Fig. \ref{subfig:mu1nu1_multi_track}.  Such a desired behavior consists of two clusters of agents orbiting on a circle. Results are reported in Fig. \ref{subfig:dkl_multi_track}, where the time evolution $\bar{E}$ is also shown. After an initial transient, $\bar{E}$ settles to approximately 50\%. For brevity, we omit the KL divergence in time, which remains below 0.3.

\subsection{Results and Discussion}\label{sec:discussion}
We considered a hybrid swarm of 4 differential drive robots and 96 virtual agents, interacting through a repulsive kernel. Assuming the group to start on a perfect square lattice (intial constant density), we tasked the swarm to aggregate according to four different desired densities, under a new 2D continuification control action. Specifically, we presented a monomodal and multimodal regulation case, where the means of the von Mises functions to achieve are time invariant, and a monomodal and multimodal tracking case, where, instead, the means of the von Mises functions to achieve are time variant. 

We characterized the performance of each trial using the time evolution of the normalized $\mathcal{L}^2$ error, namely $\bar{E}$. Although the correct formation has been attained in each of the trials, we obtained our best results in the regulations scenarios (monomodal and multimodal), where the steady-state residual percentage error went below 10\% and 30\% respectively (Fig.s \ref{fig:monomodal_regulation} and \ref{fig:multimodal_regulation}). Concerning the tracking cases, instead, performance was less remarkable, with $\bar{E}$ being around 50\%, in both the monomodal and multimodal case (see Fig. \ref{fig:tracking}). 

Although the prescribed formation was always attained (see Figs. \ref{fig:monomodal_regulation} and \ref{fig:multimodal_regulation} and available videos for the tracking cases), the asymptotic convergence that is prescribed by the theory (see Section \ref{sec:theoreticalframework}) was not accomplished. This is due to two main factors. First, we adapted the theoretical framework to experiments to cope with the periodicity assumption about the domain and with the constrained kinematic of the differential robots. Second, the inherent uncertainties and noise of the experimental set-up need also to be considered. Note that, another source of performance degradation is the finite size of the swarm. Specifically, our convergence guarantees hold in the limiting case of infinite agents. Indeed, should we numerically integrate \eqref{eq:d_dimensional_macro}, assuming an infinite number of agents, we would be able to reduce $\Vert e^F \Vert_2^2$ to 0. 


\section{Conclusions}\label{sec:conclusions}
We developed a new mixed reality, flexible, experimental environment for large-scale swarm robotics experiments with relatively small time and resources demand, and we presented the extension to higher dimensions of the continuification-based control strategy proposed in \cite{maffettone2022continuification}. Our approach leveraged hybrid swarms of differential drive robots and virtual agents, making the size of the swarm easily scalable by the user. We demonstrated the applicability and effectiveness of our set-up for the experimental validation of the continuification-based control of swarming robots in the plane. 

When experimentally implementing a macroscopic control technique with the assumption of an infinite number of agents, we reported a performance degradation, even if convergence is theoretically ensured. This is due to both implementation problems and theoretical drawbacks of the strategy. In particular, performance degradation is due to ($i$) the experimental set-up,  ($ii$) the necessary adaptation of the control strategy to the kinematic constraints of the real robots and the periodicity of the domain, and ($iii$) the inherent approximation introduced by the continuum hypothesis.
Current work seeks to build more differential drive robots to asses how the ratio between real robots and virtual agents influence the effectiveness of the platform, and rephrase the theoretical framework to reduce the number of adaptations to go from theory and simulations to reality.

\bibliographystyle{IEEEtran}

\begin{thebibliography}{10}
\providecommand{\url}[1]{#1}
\csname url@samestyle\endcsname
\providecommand{\newblock}{\relax}
\providecommand{\bibinfo}[2]{#2}
\providecommand{\BIBentrySTDinterwordspacing}{\spaceskip=0pt\relax}
\providecommand{\BIBentryALTinterwordstretchfactor}{4}
\providecommand{\BIBentryALTinterwordspacing}{\spaceskip=\fontdimen2\font plus
\BIBentryALTinterwordstretchfactor\fontdimen3\font minus
  \fontdimen4\font\relax}
\providecommand{\BIBforeignlanguage}[2]{{%
\expandafter\ifx\csname l@#1\endcsname\relax
\typeout{** WARNING: IEEEtran.bst: No hyphenation pattern has been}%
\typeout{** loaded for the language `#1'. Using the pattern for}%
\typeout{** the default language instead.}%
\else
\language=\csname l@#1\endcsname
\fi
#2}}
\providecommand{\BIBdecl}{\relax}
\BIBdecl

\bibitem{maffettone2022continuification}
G.~C. Maffettone, A.~Boldini, M.~Di~Bernardo, and M.~Porfiri,
  ``Continuification control of large-scale multiagent systems in a ring,''
  \emph{IEEE Control Systems Letters}, vol.~7, pp. 841--846, 2023.

\bibitem{maffettone2023continuification}
G.~C. Maffettone, M.~Porfiri, and M.~Di~Bernardo, ``Continuification control of
  large-scale multiagent systems under limited sensing and structural
  perturbations,'' \emph{IEEE Control Systems Letters}, vol.~7, pp. 2425--2430,
  2023.

\bibitem{nikitin2021continuation}
D.~Nikitin, C.~Canudas-de Wit, and P.~Frasca, ``A continuation method for
  large-scale modeling and control: From {ODE}s to {PDE}, a round trip,''
  \emph{IEEE Transactions on Automatic Control}, vol.~67, no.~10, pp.
  5118--5133, 2022.

\bibitem{gao2019graphon}
S.~Gao and P.~E. Caines, ``Graphon control of large-scale networks of linear
  systems,'' \emph{IEEE Transactions on Automatic Control}, vol.~65, no.~10,
  pp. 4090--4105, 2019.

\bibitem{gao2023lqg}
S.~Gao, P.~E. Caines, and M.~Huang, ``{LQG} graphon mean field games: Analysis
  via graphon invariant subspaces,'' \emph{IEEE Transactions on Automatic
  Control}, vol.~68, no.~12, pp. 7482--7497, 2023.

\bibitem{bernoff2011primer}
A.~J. Bernoff and C.~M. Topaz, ``A primer of swarm equilibria,'' \emph{SIAM
  Journal on Applied Dynamical Systems}, vol.~10, no.~1, pp. 212--250, 2011.

\bibitem{sinigaglia2022density}
C.~Sinigaglia, A.~Manzoni, and F.~Braghin, ``Density control of large-scale
  particles swarm through {PDE}-constrained optimization,'' \emph{IEEE
  Transactions on Robotics}, vol.~38, no.~6, pp. 3530--3549, 2022.

\bibitem{diBernardo2020}
M.~di~Bernardo, ``Controlling collective behavior in complex systems,'' in
  \emph{Encyclopedia of Systems and Control}, J.~Baillieul and T.~Samad,
  Eds.\hskip 1em plus 0.5em minus 0.4em\relax Springer London, 2020.

\bibitem{freudenthaler2020pde}
G.~Freudenthaler and T.~Meurer, ``{PDE}-based multi-agent formation control
  using flatness and backstepping: Analysis, design and robot experiments,''
  \emph{Automatica}, vol. 115, p. 108897, 2020.

\bibitem{biswal2021decentralized}
S.~Biswal, K.~Elamvazhuthi, and S.~Berman, ``Decentralized control of
  multiagent systems using local density feedback,'' \emph{IEEE Transactions on
  Automatic Control}, vol.~67, no.~8, pp. 3920--3932, 2021.

\bibitem{Karafyllis2022}
I.~Karafyllis, D.~Theodosis, and M.~Papageorgiou, ``{Analysis and control of a
  non-local PDE traffic flow model},'' \emph{International Journal of Control},
  vol.~95, no.~3, pp. 660--678, 2022.

\bibitem{blandin2011general}
S.~Blandin, D.~Work, P.~Goatin, B.~Piccoli, and A.~Bayen, ``A general phase
  transition model for vehicular traffic,'' \emph{SIAM journal on Applied
  Mathematics}, vol.~71, no.~1, pp. 107--127, 2011.

\bibitem{rubio2022open}
A.~Rubio~Denniss, T.~E. Gorochowski, and S.~Hauert, ``An open platform for
  high-resolution light-based control of microscopic collectives,''
  \emph{Advanced Intelligent Systems}, p. 2200009, 2022.

\bibitem{calabrese2021spontaneous}
C.~Calabrese, M.~Lombardi, E.~Bollt, P.~De~Lellis, B.~G. Bardy, and
  M.~Di~Bernardo, ``Spontaneous emergence of leadership patterns drives
  synchronization in complex human networks,'' \emph{Scientific Reports},
  vol.~11, no.~1, pp. 1--12, 2021.

\bibitem{dorigo2021swarm}
M.~Dorigo, G.~Theraulaz, and V.~Trianni, ``Swarm robotics: Past, present, and
  future [point of view],'' \emph{Proceedings of the IEEE}, vol. 109, no.~7,
  pp. 1152--1165, 2021.

\bibitem{d2023controlling}
R.~M. D’Souza, M.~di~Bernardo, and Y.-Y. Liu, ``Controlling complex networks
  with complex nodes,'' \emph{Nature Reviews Physics}, vol.~5, no.~4, pp.
  250--262, 2023.

\bibitem{slavkov2018morphogenesis}
I.~Slavkov, D.~Carrillo-Zapata, N.~Carranza, X.~Diego, F.~Jansson, J.~Kaandorp,
  S.~Hauert, and J.~Sharpe, ``Morphogenesis in robot swarms,'' \emph{Science
  Robotics}, vol.~3, no.~25, p. 9178, 2018.

\bibitem{rubenstein2013collective}
M.~Rubenstein, A.~Cabrera, J.~Werfel, G.~Habibi, J.~McLurkin, and R.~Nagpal,
  ``Collective transport of complex objects by simple robots: theory and
  experiments,'' in \emph{Proceedings of the 2013 international conference on
  Autonomous agents and multi-agent systems}, 2013, pp. 47--54.

\bibitem{caprari2005mobile}
G.~Caprari and R.~Siegwart, ``Mobile micro-robots ready to use: Alice,'' in
  \emph{2005 IEEE/RSJ International Conference on Intelligent Robots and
  Systems}.\hskip 1em plus 0.5em minus 0.4em\relax IEEE, 2005, pp. 3295--3300.

\bibitem{mondada2009puck}
F.~Mondada, M.~Bonani, X.~Raemy, J.~Pugh, C.~Cianci, A.~Klaptocz, S.~Magnenat,
  J.-C. Zufferey, D.~Floreano, and A.~Martinoli, ``The e-puck, a robot designed
  for education in engineering,'' in \emph{Proceedings of the 9th conference on
  autonomous robot systems and competitions}, vol.~1, 2009, pp. 59--65.

\bibitem{durham2011discrete}
J.~W. Durham, R.~Carli, P.~Frasca, and F.~Bullo, ``Discrete partitioning and
  coverage control for gossiping robots,'' \emph{IEEE Transactions on
  Robotics}, vol.~28, no.~2, pp. 364--378, 2011.

\bibitem{manas2023scalability}
F.~J. Ma{\~n}as-{\'A}lvarez, M.~Guinaldo, R.~Dormido, and S.~Dormido-Canto,
  ``Scalability of cyber-physical systems with real and virtual robots in ros
  2,'' \emph{Sensors}, vol.~23, no.~13, p. 6073, 2023.

\bibitem{karunarathna2023mixed}
D.~Karunarathna, N.~Jaliyagoda, G.~Jayalath, J.~Alawatugoda, R.~Ragel, and
  I.~Nawinne, ``Mixed-reality based multi-agent robotics framework for
  artificial swarm intelligence experiments,'' \emph{IEEE Access}, 2023.

\bibitem{boldini2021virtual}
A.~Boldini, X.~Ma, J.-R. Rizzo, and M.~Porfiri, ``A virtual reality interface
  to test wearable electronic travel aids for the visually impaired,'' in
  \emph{Nano-, Bio-, Info-Tech Sensors and Wearable Systems}, vol. 11590.\hskip
  1em plus 0.5em minus 0.4em\relax SPIE, 2021, pp. 50--56.

\bibitem{ricci2023virtual}
F.~S. Ricci, A.~Boldini, X.~Ma, M.~Beheshti, D.~R. Geruschat, W.~H. Seiple,
  J.-R. Rizzo, and M.~Porfiri, ``Virtual reality as a means to explore
  assistive technologies for the visually impaired,'' \emph{PLOS Digital
  Health}, vol.~2, no.~6, p. 275, 2023.

\bibitem{naik2020animals}
H.~Naik, R.~Bastien, N.~Navab, and I.~D. Couzin, ``Animals in virtual
  environments,'' \emph{IEEE Transactions on Visualization and Computer
  Graphics}, vol.~26, no.~5, pp. 2073--2083, 2020.

\bibitem{stowers2017virtual}
J.~R. Stowers, M.~Hofbauer, R.~Bastien, J.~Griessner, P.~Higgins, S.~Farooqui,
  R.~M. Fischer, K.~Nowikovsky, W.~Haubensak, I.~D. Couzin \emph{et~al.},
  ``Virtual reality for freely moving animals,'' \emph{Nature methods},
  vol.~14, no.~10, pp. 995--1002, 2017.

\bibitem{karakaya2020behavioral}
M.~Karakaya, S.~Macr{\`\i}, and M.~Porfiri, ``Behavioral teleporting of
  individual ethograms onto inanimate robots: experiments on social
  interactions in live zebrafish,'' \emph{iScience}, vol.~23, no.~8, 2020.

\bibitem{polverino2022ecology}
G.~Polverino, V.~R. Soman, M.~Karakaya, C.~Gasparini, J.~P. Evans, and
  M.~Porfiri, ``Ecology of fear in highly invasive fish revealed by robots,''
  \emph{iScience}, vol.~25, no.~1, 2022.

\bibitem{pickem2017robotarium}
D.~Pickem, P.~Glotfelter, L.~Wang, M.~Mote, A.~Ames, E.~Feron, and
  M.~Egerstedt, ``The robotarium: A remotely accessible swarm robotics research
  testbed,'' in \emph{2017 IEEE International Conference on Robotics and
  Automation (ICRA)}.\hskip 1em plus 0.5em minus 0.4em\relax IEEE, 2017, pp.
  1699--1706.

\bibitem{reina2017ark}
A.~Reina, A.~J. Cope, E.~Nikolaidis, J.~A. Marshall, and C.~Sabo, ``Ark:
  Augmented reality for kilobots,'' \emph{IEEE Robotics and Automation
  letters}, vol.~2, no.~3, pp. 1755--1761, 2017.

\bibitem{feola2023multi}
L.~Feola, A.~Reina, M.~S. Talamali, and V.~Trianni, ``Multi-swarm interaction
  through augmented reality for kilobots,'' \emph{IEEE Robotics and Automation
  Letters}, 2023.

\bibitem{viscek1995}
T.~Viscek, A.~Czir{\`{o}}k, E.~Ben-Jacob, I.~Cohen, and O.~Shochet, ``{Novel
  Type of Phase Transition in a System of Self-Driven Particles},''
  \emph{Physical Review Letters}, vol.~75, no.~6, pp. 1226--1229, 1995.

\bibitem{siciliano}
B.~Siciliano, L.~Sciavicco, L.~Villani, and G.~Oriolo, \emph{Robotics:
  Modelling, Planning and Control}.\hskip 1em plus 0.5em minus 0.4em\relax
  Springer Publishing Company, Incorporated, 2010.

\bibitem{jeruchim2006simulation}
M.~C. Jeruchim, P.~Balaban, and K.~S. Shanmugan, \emph{Simulation of
  communication systems: modeling, methodology and techniques}.\hskip 1em plus
  0.5em minus 0.4em\relax Springer Science \& Business Media, 2006.

\bibitem{Dorsogna2006}
M.~R. D'Orsogna, Y.~L. Chuang, A.~L. Bertozzi, and L.~S. Chayes,
  ``{Self-propelled particles with soft-core interactions: Patterns, stability,
  and collapse},'' \emph{Physical Review Letters}, vol.~96, no.~10, 2006.

\bibitem{griffiths2005introduction}
D.~J. Griffiths, ``Introduction to electrodynamics,'' 2005.

\bibitem{maffettone2024high}
G.~C. Maffettone, M.~di~Bernardo, and M.~Porfiri, ``High-dimensional
  continuification control of large-scale multi-agent systems under limited
  sensing and perturbations,'' \emph{arXiv preprint arXiv:2403.17191}, 2024.

\bibitem{stein2011fourier}
E.~M. Stein and R.~Shakarchi, \emph{Fourier analysis: an introduction}.\hskip
  1em plus 0.5em minus 0.4em\relax Princeton University Press, 2011, vol.~1.

\bibitem{ando1999distributed}
H.~Ando, Y.~Oasa, I.~Suzuki, and M.~Yamashita, ``Distributed memoryless point
  convergence algorithm for mobile robots with limited visibility,'' \emph{IEEE
  Transactions on Robotics and Automation}, vol.~15, no.~5, pp. 818--828, 1999.

\bibitem{oh2015survey}
K.-K. Oh, M.-C. Park, and H.-S. Ahn, ``A survey of multi-agent formation
  control,'' \emph{Automatica}, vol.~53, pp. 424--440, 2015.

\bibitem{gagliardi2022probabilistic}
D.~Gagliardi and G.~Russo, ``On a probabilistic approach to synthesize control
  policies from example datasets,'' \emph{Automatica}, vol. 137, p. 110121,
  2022.

\end{thebibliography}
\end{document}